\makeatletter\def\graphicscache@inhibit{true}\makeatother
\documentclass[iicol]{sn-jnl}%

\usepackage[utf8]{inputenc}
\usepackage[style=nature,hyperref,natbib=true,backend=biber,maxbibnames=99]{biblatex}
\bibliography{references.bib}

\usepackage{manyfoot}

\usepackage{url}
\usepackage{graphicx}
\usepackage[draft,inline, nomargin]{fixme}
\fxsetup{theme=color}

\usepackage{amsmath}
\usepackage{amssymb}
\usepackage{textcomp}
\usepackage{siunitx}

\usepackage{booktabs}
\usepackage{threeparttable}
\usepackage{multirow}

\usepackage[capitalize]{cleveref}
\crefformat{section}{Sec.~#2#1#3}
\crefformat{table}{Tab.~#2#1#3}

\usepackage{tikz}
\usetikzlibrary{arrows}
\usetikzlibrary{positioning,calc}
\usetikzlibrary{decorations.pathreplacing}
\usetikzlibrary{decorations.markings}
\usetikzlibrary{fit}
\usetikzlibrary{shapes.callouts}
\usetikzlibrary{shapes.geometric}
\usetikzlibrary{matrix}
\usetikzlibrary{arrows}
\usetikzlibrary{decorations.pathmorphing}
\usetikzlibrary{calligraphy}
\usetikzlibrary{shapes.arrows}
\usetikzlibrary{spy}
\usepackage{setspace}

\pgfdeclarelayer{background}
\pgfdeclarelayer{foreground}
\pgfsetlayers{background,main,foreground}

\usepackage{scalefnt}

\usepackage{pgfplots}
\pgfplotsset{compat=1.14}
\usepgfplotslibrary{groupplots}
\usepgfplotslibrary{units}
\usepgfplotslibrary{statistics}
\usepgfplotslibrary{colorbrewer}

\usepackage{pgfplotstable}

\tikzset{
vr/.style={fill=Pastel2-A},
manip/.style={fill=Pastel2-B},
loco/.style={fill=Pastel2-D},
system/.style={fill=Pastel2-C},
}

\usepackage[export]{adjustbox}

\definecolor{teaserbg}{HTML}{ebebeb}

\usepackage{ifthen}
\usepackage{graphicscache}

\usepackage[tightpage]{preview}

\newenvironment{maybepreview}%
{\noindent\ignorespaces}%
{\par\noindent%
\ignorespacesafterend}
\PreviewEnvironment{maybepreview}

\tikzset{
  font=\sffamily\footnotesize,
  m/.style={draw, rounded corners, fill=yellow!20, align=center}
}

\definecolor{t1}{RGB}{235, 172, 35}
\definecolor{t2}{RGB}{184, 0, 88}
\definecolor{t3}{RGB}{0, 140, 249}
\definecolor{t4}{RGB}{0, 110, 0}
\definecolor{t5}{RGB}{0, 187, 173}
\definecolor{t6}{RGB}{209, 99, 230}
\definecolor{t7}{RGB}{178, 69, 2}
\definecolor{t8}{RGB}{255, 146, 135}
\definecolor{t9}{RGB}{89, 84, 214}
\definecolor{t10}{RGB}{135, 133, 0}
\definecolor{t11}{RGB}{0, 198, 248}
\definecolor{t12}{RGB}{0, 167, 108}
\definecolor{t13}{RGB}{189, 189, 189}

\tikzdeclarecoordinatesystem{rel}{%
	\tikzset{cs/.cd,x=0pt,y=0pt,#1}%
	\pgfpointlineattime{(\tikz@cs@x)}%
	{\pgfpointanchor{\tikz@pp@name{\tikz@cs@node}}{south west}}%
	{\pgfpointanchor{\tikz@pp@name{\tikz@cs@node}}{south east}}%
	\edef\tikz@cs@x{\the\pgf@x}%
	\pgfpointlineattime{(\tikz@cs@y)}%
	{\pgfpointanchor{\tikz@pp@name{\tikz@cs@node}}{south west}}%
	{\pgfpointanchor{\tikz@pp@name{\tikz@cs@node}}{north west}}%
	\pgfpoint{\tikz@cs@x}{\pgf@y}%
}

\tikzdeclarecoordinatesystem{rel2}{%
	\tikzset{cs/.cd,x=0pt,y=0pt,#1}%
	\pgfpointlineattime{(\tikz@cs@x / 100)}%
	{\pgfpointanchor{\tikz@pp@name{\tikz@cs@node}}{south west}}%
	{\pgfpointanchor{\tikz@pp@name{\tikz@cs@node}}{south east}}%
	\edef\tikz@cs@x{\the\pgf@x}%
	\pgfpointlineattime{(\tikz@cs@y / 100)}%
	{\pgfpointanchor{\tikz@pp@name{\tikz@cs@node}}{south west}}%
	{\pgfpointanchor{\tikz@pp@name{\tikz@cs@node}}{north west}}%
	\pgfpoint{\tikz@cs@x}{\pgf@y}%
}

\newcommand\currentcoordinate{\the\tikz@lastxsaved,\the\tikz@lastysaved}

\pgfplotsset{
    boxplot discard if not/.style 2 args={
        /pgfplots/boxplot/data filter/.code={
            \edef\tempa{\thisrow{#1}}
            \edef\tempb{#2}
            \ifx\tempa\tempb
            \else
                
            \fi
        }
    }
}
\pgfplotsset{
    discard if not/.style 2 args={
        x filter/.code={
            \edef\tempa{\thisrow{#1}}
            \edef\tempb{#2}
            \ifx\tempa\tempb
            \else
                
            \fi
        }
    }
}
\pgfplotsset{filter discard warning=false}
\makeatletter
\pgfplotsset{
    boxplot/hide outliers/.code={
        \def\pgfplotsplothandlerboxplot@outlier{}%
    }
}
\makeatother

\newsavebox\CBox 
\newcommand{\textbb}[1]{\sbox\CBox{#1}\resizebox{\wd\CBox}{\ht\CBox}{\textbf{#1}}}

\DeclareRobustCommand{\taskb}[1]{\strut\tikz{\node[fill=#1,rounded corners=1pt,minimum height=1.5ex,minimum width=1.5ex]{};}}

\begin{document}

\title[NimbRo Avatar]{NimbRo wins ANA Avatar XPRIZE Immersive Telepresence Competition: Human-Centric Evaluation and Lessons Learned}

\author*[1]{\fnm{Christian} \sur{Lenz}}\email{lenz@ais.uni-bonn.de}
\equalcont{These authors contributed equally to this work.}
\author[1]{\fnm{Max} \sur{Schwarz}}\email{schwarz@ais.uni-bonn.de}
\equalcont{These authors contributed equally to this work.}

\author[1]{\fnm{Andre} \sur{Rochow}}\email{rochow@ais.uni-bonn.de}
\author[1]{\fnm{Bastian} \sur{P\"atzold}}\email{paetzold@ais.uni-bonn.de}
\author[1]{\fnm{Raphael} \sur{Memmesheimer}}\email{memmesheimer@ais.uni-bonn.de}
\author[1]{\fnm{Michael} \sur{Schreiber}}\email{schreiber@ais.uni-bonn.de}
\author[1]{\fnm{Sven} \sur{Behnke}}\email{behnke@cs.uni-bonn.de}

\affil[1]{\orgdiv{Autonomous Intelligent Systems}, \orgname{University of Bonn}, \country{Germany}}

\abstract{
Robotic avatar systems can enable immersive telepresence with locomotion, manipulation,
and communication capabilities.
We present such an avatar system, based on the key components of immersive
3D visualization and transparent force-feedback telemanipulation.
Our avatar robot features an anthropomorphic upper body with
dexterous hands. The remote human operator drives the arms and fingers through
an exoskeleton-based operator station, which provides force feedback both
at the wrist and for each finger.
The robot torso is mounted on a holonomic base, providing omnidirectional locomotion on flat floors, controlled using a 3D rudder device.
Finally, the robot features a 6D movable head with stereo cameras, which stream
images to a VR display worn by the operator. Movement latency is hidden using spherical
rendering. The head also carries a telepresence screen displaying an animated image
of the operator's face, enabling direct interaction with remote persons.
Our system won the \$10M ANA Avatar XPRIZE competition, which challenged teams to develop intuitive and immersive avatar systems that could be operated by briefly trained judges.
We analyze our successful participation in the semifinals and finals and provide insight into our operator training and lessons learned.
In addition, we evaluate our system in a user study that demonstrates its intuitive and easy usability.
}

\keywords{Telepresence, Avatar, Dual-Arm Teleoperation, Face Animation, Competitions}

\maketitle

\DeclareRobustCommand{\compb}[2][system]{\strut\tikz[baseline]{\node[draw,#1,anchor=base,rounded corners,minimum height=1.5ex,minimum width=1.5ex,text depth=0pt]{#2};}}
\begin{figure*}
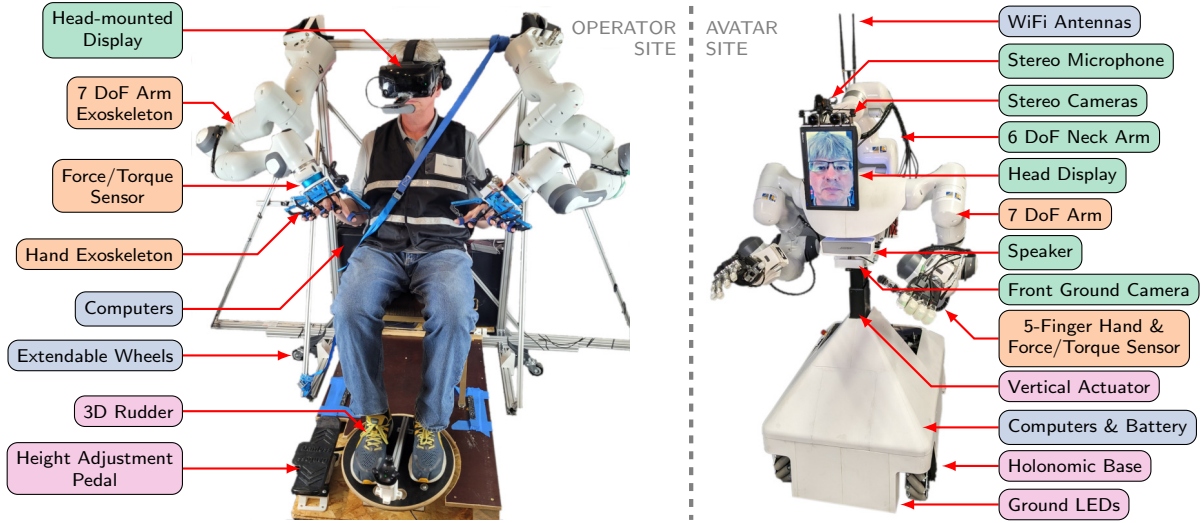

\centering\begin{maybepreview}%
\tikzset{
    n/.style={fill=yellow!20,rounded corners,draw=black,text=black,align=center,thin,font=\sffamily\footnotesize},
    lab/.style={draw=red,latex-,thick}
 }
 \centering
 \tikz[]{
  \node[inner sep=0pt] (otto) {\includegraphics[height=6.8cm]{images/jerry.jpg}};

  \coordinate (ottow) at ($(otto.west)+(-0.1,0)$);

  \draw[lab] (rel cs:x=0.48,y=0.88,name=otto) -- ++(135:0.7cm) -- (\currentcoordinate-|ottow) node [n,vr,anchor=east] {Head-mounted\\Display};
  \draw[lab] (rel cs:x=0.1,y=0.77,name=otto) -- ++(135:0.4cm) -- (\currentcoordinate-|ottow) node [n,manip,anchor=east] {7 DoF Arm\\Exoskeleton};
  \draw[lab] (rel cs:x=0.25,y=0.65,name=otto) --  (\currentcoordinate-|ottow) node [n,manip,anchor=east] {Force/Torque\\Sensor};
  \draw[lab] (rel cs:x=0.25,y=0.6,name=otto) -- ++(225:0.8cm) -- (\currentcoordinate-|ottow) node [n,manip,anchor=east] {Hand Exoskeleton};

  \draw[lab] (rel cs:x=0.35,y=0.535,name=otto) -- ++(225:1.2cm) -- (\currentcoordinate-|ottow) node [n,system,anchor=east] {Computers};

  \draw[lab] (rel cs:x=0.23,y=0.32,name=otto) -- (\currentcoordinate-|ottow) node [n,system,anchor=east] {Extendable Wheels};

  \draw[lab] (rel cs:x=0.4,y=0.17,name=otto) -- ++(135:0.4cm) -- (\currentcoordinate-|ottow) node [n,loco,anchor=east] {3D Rudder};
  \draw[lab] (rel cs:x=0.25,y=0.1,name=otto) -- (\currentcoordinate-|ottow) node [n,loco,anchor=east] {Height Adjustment\\Pedal};

  \node[inner sep=0pt,right=1cm of otto] (anna) {\includegraphics[height=6.8cm]{images/anna.jpg}};

  \coordinate (annae) at ($(anna.east)+(0.2,0)$);

  \draw[lab] (rel cs:x=0.55,y=0.97,name=anna) -- (\currentcoordinate-|annae) node [n,system,anchor=west] {WiFi Antennas};

  \draw[lab] (rel cs:x=0.45,y=0.82,name=anna) -- ++(45:0.7cm) -- (\currentcoordinate-|annae) node [n,vr,anchor=west] {Stereo Microphone};
  \draw[lab] (rel cs:x=0.53,y=0.79,name=anna) -- ++(45:0.25cm) -- (\currentcoordinate-|annae) node [n,vr,anchor=west] {Stereo Cameras};

  \draw[lab] (rel cs:x=0.7,y=0.745,name=anna) -- (\currentcoordinate-|annae) node [n,vr,anchor=west] {6 DoF Neck Arm};
  \draw[lab] (rel cs:x=0.55,y=0.67,name=anna) -- (\currentcoordinate-|annae) node [n,vr,anchor=west] {Head Display};
  \draw[lab] (rel cs:x=0.9,y=0.595,name=anna) -- (\currentcoordinate-|annae) node [n,manip,anchor=west] {7 DoF Arm};

  \draw[lab] (rel cs:x=0.6,y=0.52,name=anna) -- (\currentcoordinate-|annae) node [n,vr,anchor=west] {Speaker};
  \draw[lab] (rel cs:x=0.55,y=0.495,name=anna) -- ++(-45:0.45cm) -- (\currentcoordinate-|annae) node [n,vr,anchor=west] {Front Ground Camera};

  \draw[lab] (rel cs:x=0.86,y=0.41,name=anna) -- ++ (-45:0.55cm) -- (\currentcoordinate-|annae) node [n,manip,anchor=west] {5-Finger Hand \&\\Force/Torque Sensor};

  \draw[lab] (rel cs:x=0.55,y=0.40,name=anna) -- ++(-45:1.35cm) -- (\currentcoordinate-|annae) node [n,loco,anchor=west] {Vertical Actuator};

  \draw[lab] (rel cs:x=0.78,y=0.18,name=anna) -- (\currentcoordinate-|annae) node [n,system,anchor=west] {Computers \& Battery};

  \draw[lab] (rel cs:x=0.85,y=0.105,name=anna) -- (\currentcoordinate-|annae) node [n,loco,anchor=west] {Holonomic Base};

  \draw[lab] (rel cs:x=0.7,y=0.03,name=anna) -- (\currentcoordinate-|annae) node [n,loco,anchor=west] {Ground LEDs};

  \coordinate (split) at ($(otto.north east)!0.8!(anna.north west)$);
  \begin{scope}[draw=black!50,text=black!50]
    \node[below right=0.1cm of split,align=left] {AVATAR\\SITE};
    \node[below left=0.1cm of split,align=right] {OPERATOR\\SITE};
    \draw[dashed,ultra thick] (split) -- ($(otto.south east)!0.8!(anna.south west)$);
  \end{scope}
 }
  \end{maybepreview}%
 \caption{NimbRo avatar system consisting of the operator station (left) and the avatar robot (right).
 Colors correspond to categories (\compb[vr]{Audiovisual telepresence}, \compb[manip]{Telemanipulation}, \compb[loco]{Locomotion}, \compb[system]{System / Misc}).}
 \label{fig:sys_overview}
\end{figure*}

\section{Introduction}
\begin{tikzpicture}[remember picture,overlay]
  \node[anchor=north,align=center,font=\sffamily\small,yshift=-0.4cm] at (current page.north) {%
  \textbf{Accepted for International Journal of Social Robotics (SORO), Springer, to appear 2023.}%
  };%
\end{tikzpicture}%
Robotic avatar systems combine high-quality telecommunication
with intuitive robotic teleoperation, creating true telepresence.
These systems allow full immersion into a remote space while also \textit{embodying} the
operator in a robotic system, giving them the ability to navigate in the remote environment,
to manipulate objects there, and to interact with remote persons in a multimodal way that includes direct physical contact.

Through such an immersive telepresence system, humans can perform tasks in remote environments which are currently beyond the capabilities of autonomous
perception, planning, and control methods---the human intellect is still unmatched in its ability to perceive, plan, and react to unforeseen situations.
Avatar systems allow humans to work in remote environments without having to travel
or expose themselves to potential dangers, such as in disaster response.

The ANA Avatar XPRIZE competition\footnote{\url{https://www.xprize.org/prizes/avatar}} challenged the robotics community
to advance the state of the art in immersive telepresence systems.
Equipped with a record \$10M prize purse, the competition required teams to build \textit{intuitive} and \textit{robust} robotic avatar systems that allow a human operator to be present in a remote space.
The tasks to be solved included social interaction and communication, but also locomotion and complex manipulation.
Critically, the systems were to be used and evaluated by operator and recipient judges.
In contrast to previous teleoperation competitions such as the DARPA Robotics Challenge~\citep{krotkov2018darpa}, operators could be trained only for a short time to use the developed avatar systems.

Our NimbRo Avatar system (\cref{fig:sys_overview}) won the ANA Avatar XPRIZE Finals in November 2022\footnote{\url{https://www.youtube.com/watch?v=EmESa2Olq4c}}.
This article aims to provide a comprehensive overview of our system by summarizing previous publications \citep{schwarz2021nimbro,schwarz2021vr,lenz2023bimanual,schwarz2023robust,paetzold2023haptics,rochow2023attention}.
It also extends these previous works by focusing on the human interaction, both between operator and machine as well as
operator and human recipients interacting with the avatar. Our contributions include
\begin{itemize}
 \item an avatar robot with humanoid upper body and the accompanying operator station,
 \item bidirectional audiovisual telepresence including latency-free rendering during head movement
   as well as operator face animation,
 \item transparent dual-arm telemanipulation with force feedback on arm and finger level,
 \item roughness sensing and haptic display,
 \item safety, monitoring, and robustness modules,
 \item a detailed analysis of our competition performance as well as a user study, and
 \item a discussion of lessons learned.
\end{itemize}

\section{Related Work}

Telemanipulation robots are complex systems consisting of many components, which have been investigated both individually as well as on a systems level.

\paragraph{Telemanipulation Systems}

The DARPA Robotics Challenge (DRC) 2015~\citep{krotkov2018darpa} resulted in the development of several
mobile telemanipulation robots, such as DRC-HUBO~\citep{oh2017technical}, CHIMP~\citep{stentz2015chimp}, RoboSimian~\citep{karumanchi2017team}, and our own entry Momaro~\citep{schwarz2017nimbro}.
All these systems demonstrated impressive locomotion and manipulation capabilities under teleoperation,
even with severely constrained communication.
The DRC placed no emphasis on intuitiveness of the teleoperation controls or on immersion of
the operators, though.
To our knowledge, our team was the only one using a VR head-mounted display (HMD) and 6D magnetic trackers
to perceive the environment in 3D and to control the robot arms~\citep{rodehutskors2015intuitive}---all other teams relying entirely on 2D monitors and traditional input devices to control their robots.
All teams, including ours, required highly trained operators familiar with the custom-designed
operator interfaces.
Furthermore, since the DRC was geared towards disaster response, the robots did not feature any
communication capabilities for interacting with remote humans.

In our subsequent work~\citep{klamt2020remote}, we developed the ideas embodied in the Momaro system further.
The resulting Centauro robot is a torque-controlled platform capable of locomotion and dexterous manipulation in rough
terrain. It is controlled by a human sitting in a dedicated operator station, equipped with
an upper-body exoskeleton providing force feedback and a VR HMD.
Still, Centauro is focused on disaster response and does not have any communication facilities.

Recently, there has been explosive growth in teleoperated humanoid robots.
\citet{darvish2023teleoperation} provide a comprehensive overview comparing fully-humanoid (i.e. walking)
robotic teleoperation systems.
Walking humanoid robots can overcome obstacles, rough terrain, and stairs, but raise more complex challenges regarding balance and whole-body control.
As the XPRIZE competition as well as many human-made environments feature flat surfaces on which wheels are more efficient and stable, our system features a wheeled omnidirectional base.

\Citet{Schmaus:RAL18} discuss the results of the METRON SUPVIS Justin space-robotics experiment,
where an astronaut on the international space station controlled the Justin robot on Earth, simulating an orbital robotics
mission. Instead of opting for full immersion and direct control, the authors relied on a 2D tablet
display and higher levels of autonomy, allowing the astronaut to trigger autonomous task skills.
A similar approach was developed for our domestic service robot Cosero~\citep{schwarz2014mobile}.

In contrast to the discussed prior works, our avatar system is specifically designed to operate in
human workspaces and to interact with humans.
To the best of our knowledge, there were no integrated robots designed for this purpose
prior to the ANA Avatar XPRIZE, which initiated development of such systems~\citep{luo2023team,marques2022commodity,van2022comes,park2022team,lentini2019alter,vaz2022immersive,lastmile}.
Notably, \citet{luo2023team} describe the approach by the third-placed Team Northeastern, who
developed hydraulic grippers with high-fidelity force feedback. Similar to our system, the team
incorporated two Franka Emika Panda arms for bimanual manipulation on top of a omnidirectional base.
Team AVATRINA, described by \citet{marques2022commodity}, reached fourth place, again with
a bimanual manipulation system based on Franka Emika Panda arms.
In contrast to other top-placed teams, our system featured increased immersion and intuitiveness
through free 6D head movement, a photorealistic face animation system for communication with recipients,
and haptic feedback from multimodal sensors.
In addition, our team focused on system robustness, rigorous testing, and training of the crew that
later trained operators to use the avatar system.

\paragraph{3D VR Televisualization}

Live capture and visualization of the remote scene is typically done using data from RGB or \mbox{RGB-D} cameras.
There are many examples of static and movable stereo cameras on robots, which are directly visualized
in a head-mounted display~\citep{martins2015design,zhu2010head,agarwal2016imitating}. However,
these approaches are limited by either a fixed viewpoint or considerable camera movement latency,
potentially creating motion sickness. In contrast, our system hides latencies by correcting
viewpoint changes through spherical rendering~\citep{schwarz2021vr}.

RGB-D sensors allow rendering from free viewpoints~\citep{whitney2020comparing,sun2020new}, removing head movement latency. However, these sensors produce depth images with missing measurements, which can be difficult to visualize
in a convincing way. Reconstruction-based approaches~\citep{rodehutskors2015intuitive,klamt2020remote,stotko2019vr} address this issue by aggregating
3D measurements over time and building dense representations, which can be viewed without movement latency.
They still, however, struggle with many reflective and transparent materials, because the depth sensors
cannot measure them. An additional drawback is that reconstruction-based approaches usually cannot deal
with dynamic scenes---which is an issue when interacting with the environment and human recipients.
In contrast, our method always displays a live stereo RGB stream, which has no difficulties with materials
or dynamic scenes.

\paragraph{Force Feedback}

Teleoperation systems use typically stationary devices to display any force feedback captured by the
remote robot to the human operator~\cite{hirche2012human, klamt2020remote, abi2018humanoid}. In contrast, wearable haptic devices~\cite{bimbo2017teleoperation}
are usually more lightweight and do not limit the operator's workspace. However, they cannot display
absolute forces to the operator.

Much recent and ongoing research focuses on stable teleoperation systems
in time-delayed scenarios~\cite{balachandran2020closing, wang2017adaptive}. Large time
delays for teleoperation in earth-space scenarios are investigated~\cite{Panzirsch:2020, guanyang2019haptic}.
In our application, we assume smaller distances between the operator station and the avatar robot. Thus, our force feedback controller does not need to handle such high latencies.

\paragraph{Locomotion Control}

Locomotion control is a key aspect of avatar robot systems.
Directly related are locomotion interfaces for virtual reality control. 
Interesting hands-free locomotion control can be achieved by e.g. treadmills~\cite{lichtenstein2007feedback}, circular moving tiles~\cite{iwata2005circulafloor}, or walking pads~\cite{bouguila2004walking}. 
These approaches tend to be exhausting for the operator and are not suitable for long-term operation. 
Further, in our setup they might transfer unintended motion to the robot's arms.
More relaxing for the operators are seated locomotion controllers.
\Citet{ohshima2016virtual} present a device that integrates a pressure sensor in a seat cushion. 
To detect lifting of the operator's leg which is then translated to motion commands.
Some interfaces are feet-controlled. 
\Citet{carmichael2020spring} present a device that is attached to the operator's feet and is able to detect the operators foot movement.
\Citet{otaran2021short} introduced a feet controller  for linear movement by recognizing steps on the platform.
Interestingly, they integrate haptic feedback e.g. for the terrain.
Other approaches also consider leaning as input~\cite{kitson2017comparing}.

\paragraph{Facial Animation}
Visualizing facial expressions of people wearing VR HMDs is a well-known task enabling remote social
interaction.
Often, eye tracking cameras capture eye poses and expressions such as frowns, while a standard camera
captures the unobscured lower part of the face\citep{chu2020expressive,lombardi2018deep,vr_facial}.
A special requirement of the ANA Avatar XPRIZE competition was that the method had to be quickly adaptable to a new operator, as only 45\,min of setup time was allowed.

A first category of HMD facial animation methods is based on explicit 3D representations.
\Citet{olszewski2016high} train a neural regressor to output blend shape weights, which deform a face mesh.
On the other hand, Codec Avatars~\citep{chu2020expressive,lombardi2018deep,vr_facial,faceaudio} are an implicit model trained on
many images of the operator.

All the mentioned methods require either extensive manual work (3D modeling), complicated capture setups
(3D reconstruction), or long training times, all of which were infeasible in the avatar competition.
In contrast, our 2D approach is based on taking a short video of the operator and does not require
any on-site or operator-specific training. %

\begin{figure}
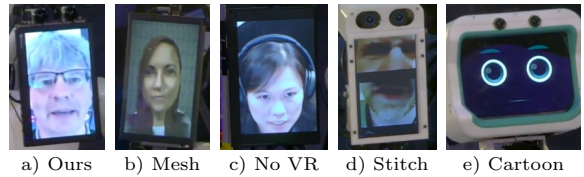
\centering\begin{maybepreview}%
	\begin{tikzpicture}[
	every label/.style={font=\footnotesize},
	every node/.style={inner sep=0pt},
	label distance=.5ex
	]
	\node[matrix, column sep=4.05pt] {
		\node[label={south:{a) Ours}}] {\includegraphics[height=1.93cm]{images/comp/nimbro.png}}; &
		\node[label={south:{b) Mesh}}] {\includegraphics[height=1.93cm]{images/comp/pollen.png}}; &
		\node[label={south:{c) No VR}}] {\includegraphics[height=1.93cm]{images/comp/northeastern.png}}; &
		\node[label={south:{d) Stitch}}] {\includegraphics[height=1.93cm]{images/comp/avatrina.png}}; &
		\node[label={south:{e) Cartoon}}] {\includegraphics[height=1.93cm,clip,trim=40px 0px 40px 0px]{images/comp/unist.png}}; \\
	};
	\end{tikzpicture}
	\end{maybepreview}%
	\caption{Types of facial animation at ANA Avatar XPRIZE finals. Examples from teams: a) NimbRo,
		b)~Pollen Robotics, c)~Northeastern~\cite{luo2022towards}, d)~AVATRINA~\citep{marques2022commodity}, e)~UNIST.}
	\label{fig:types} \vspace{-2ex}
\end{figure}

From ANA Avatar XPRIZE finals video footage we recognize five categories of face animation techniques used by participants (see \cref{fig:types}).
Our team was the only one to produce a photorealistic animated face image.

\section{NimbRo Avatar System}

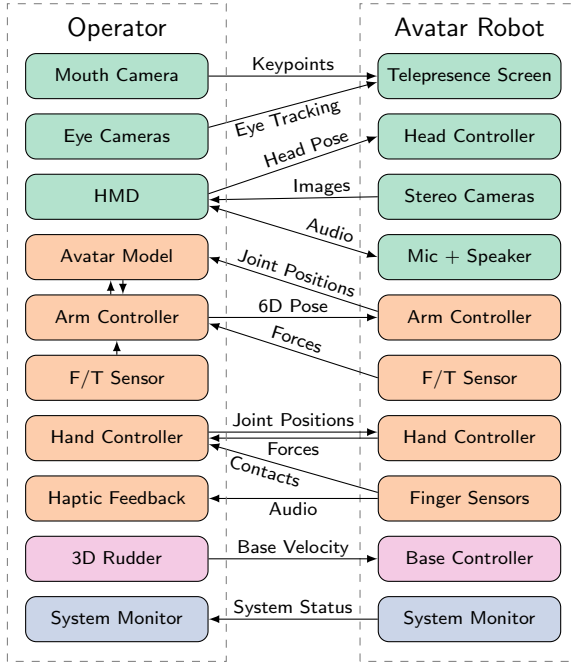
\begin{figure}
\centering\begin{maybepreview}%
\begin{tikzpicture}[
 	font=\sffamily\footnotesize,
    every node/.append style={text depth=.2ex},
	box/.style={rectangle, inner sep=0.05, anchor=west, align=center},
	line/.style={black, thick},
	midway/.append style={font=\sffamily\footnotesize},
	above/.append style={yshift=-0.5ex},
	below/.append style={yshift=0.5ex},
	scale=0.8
]
\tikzset{every node/.append style={node distance=3.0cm}}
\tikzset{terminal_node/.append style={minimum size=1.0em,minimum height=1.7em,minimum width=2.4cm,draw,align=center,rounded corners,fill=yellow!40}}
\tikzset{content_node/.append style={minimum size=1.5em,minimum height=3em,minimum width={width("Search Point")+0.2em},draw,align=center,fill=blue!15!white, rounded corners}}
\tikzset{header_node/.append style={minimum size=1.5em,minimum height=3em,minimum width=2.4cm,align=center, rounded corners}}
\tikzset{label_node/.append style={near start}}
\tikzset{group_node/.append style={align=center,rounded corners,draw, dashed , inner sep=1em,thick}}
\tikzset{decision_node/.append style={align=center,shape aspect=1.5,minimum width=7.9em,minimum height=5.4em,diamond,draw,fill=yellow!25!white,font=\sffamily\normalsize,node distance=3.9cm}}
\tikzset{l/.style={}}

\node(streamcam)[terminal_node,vr] at(0,9) {Mouth Camera};
\node(eye)[terminal_node,vr] at(0.0,8) {Eye Cameras};
\node(hmd)[terminal_node,vr] at(0.0,7) {HMD};
\node(anna_model) [terminal_node,manip] at(0.0, 6) {Avatar Model};
\node(otto_arm_controller)[terminal_node,manip] at (0.0, 5){Arm Controller};
\node(ottoFT)[terminal_node,manip] at(0.0,4) {F/T Sensor};
\node(otto_hand_controller)[terminal_node,manip] at(0.0,3) {Hand Controller};
\node(haptic_feedback)[terminal_node,manip] at(0.0,2) {Haptic Feedback};
\node(3drudder)[terminal_node,loco] at(0.0,1.0) {3D Rudder};
\node(o_sysmon)[terminal_node,system] at(0.0, 0) {System Monitor};

\begin{scope}[shift={(5.8,0.0)}]
\node(face) [terminal_node,vr] at(0.0, 9) {Telepresence Screen};
\node(head) [terminal_node,vr] at(0.0, 8) {Head Controller};
\node(cams) [terminal_node,vr] at(0.0, 7) {Stereo Cameras};
\node(annaaudio) [terminal_node,vr] at(0.0, 6) {Mic + Speaker};

\node(anna_arm_controller) [terminal_node,manip] at(0.0, 5) {Arm Controller};
\node(annaFT) [terminal_node,manip] at (0.0,4){F/T Sensor};
\node(anna_hand_controller) [terminal_node,manip] at(0.0, 3) {Hand Controller};
\node(finger_sensors) [terminal_node,manip] at(0.0, 2) {Finger Sensors};
\node(base_controller) [terminal_node,loco] at(0.0, 1.0) {Base Controller};
\node(a_sysmon) [terminal_node,system] at(0.0, 0.0) {System Monitor};

\draw [dashed, gray] (-1.8, -0.7) -- (-1.8, 10.2) -- (1.8, 10.2) -- (1.8, -0.7) -- cycle;
\node()[header_node, align = center] at (0.0, 9.8) {\normalsize Avatar Robot};
\end{scope}

\draw [l,-latex] (streamcam.east) -- (face.west) node [midway,above] {Keypoints};
\draw [l,-latex] ([yshift=0.15cm]eye.east) -- ([yshift=-0.1cm]face.west) node [pos=0.45,below,sloped] {Eye Tracking};
\draw [l,-latex] ([yshift=0.05cm]hmd.east) -- (head.west) node [pos=0.6,above,sloped] {Head Pose};
\draw [l,latex-] ([yshift=-0.05cm]hmd.east) -- (cams.west) node [pos=0.67,above,sloped] {Images};
\draw [l,latex-latex] ([yshift=-0.15cm]hmd.east) -- (annaaudio.west) node [pos=.7,above,sloped] {Audio};

\draw [l,latex-] ([yshift=-0.1cm]otto_arm_controller.east) -- (annaFT.west) node [midway,above,sloped] {Forces};
\draw [l,-latex] ([yshift=0.1cm]anna_arm_controller.west) -- (anna_model.east) node [midway,above,sloped] {Joint Positions};
\draw [l,-latex] ([xshift=0.1cm]anna_model.south) -- ([xshift=0.1cm]otto_arm_controller.north);
\draw [l,latex-] ([xshift=-0.1cm]anna_model.south) -- ([xshift=-0.1cm]otto_arm_controller.north);
\draw [l,-latex] (ottoFT.north) -- (otto_arm_controller.south);
\draw [l,-latex] (otto_arm_controller.east) -- (anna_arm_controller.west) node [midway,above,sloped] {6D Pose};

\draw [l,-latex] ([yshift=0.1cm]otto_hand_controller.east) -- ([yshift=0.1cm]anna_hand_controller.west) node [midway,above,sloped] {Joint Positions};
\draw [l,latex-] ([yshift=-0.0cm]otto_hand_controller.east) -- ([yshift=-0.0cm]anna_hand_controller.west) node [midway,below,sloped] {Forces};
\draw [l,latex-] ([yshift=-0.1cm]otto_hand_controller.east) -- ([yshift=0.1cm]finger_sensors.west) node [pos=0.35,below,sloped] {Contacts};
\draw [l,-latex] (3drudder.east) -- (base_controller.west) node [midway,above,sloped] {Base Velocity};

\draw [l,latex-] (o_sysmon.east) -- (a_sysmon.west) node [midway,above,sloped] {System Status};
\draw [l,latex-] (haptic_feedback.east) -- (finger_sensors.west) node [midway,below,sloped] {Audio};

\draw [dashed, gray] (-1.8, -0.7) -- (-1.8, 10.2) -- (1.8, 10.2) -- (1.8, -0.7) -- cycle;
\node()[header_node, align = center] at (0.0, 9.8) {\normalsize Operator};

\end{tikzpicture} 
 \end{maybepreview}%
\caption{Information flow between the system components. Same coloring as in \cref{fig:sys_overview}.}
\label{fig:system}
\end{figure}

The NimbRo avatar system consists of the operator station and the avatar robot, which allows a human operator to feel present and interact in a remote location~(\cref{fig:sys_overview,fig:system}).
The operator station includes two arm and hand exoskeletons for telemanipulation with force and haptic feedback, a Head Mounted Display (HMD) that transmits video and audio for immersive telepresence, and two foot devices for locomotion control and avatar height adjustment.
All components are connected via a standard PC (AMD Ryzen 9 5950X @ 3.4\,GHz, NVIDIA RTX A6000) and communicate via the Robot Operating System (ROS)~\citep{Quigley09} framework.
The entire operator station can be moved by extending four wheels and can be temporarily powered by an EcoFlow portable power station for about 90\,min.

The avatar robot has an anthropomorphic upper body with two arms ending in five-fingered hands, a head carrying a pair of stereo cameras, a stereo microphone, and a telepresence screen displaying the operator's face.
The upper body is attached to the mobile omnidirectional base through a height-adjustable spine.
The robot footprint is 90$\times$62\,cm and its height is between 128 and 182\,cm.
A shoulder width of 78\,cm allows navigation through narrow passages such as standard doors.
The total weight including battery and all onboard computing is approx. 140\,kg.
The robot is powered by a RELiON InSight 48\,V 30\,Ah battery, which allows approx. 2\,hours of operation.
Its base contains three
arm controllers, power supplies, and a PC (Intel i9 12900K @ 5.2\,GHz, NVIDIA RTX 3070) running a dedicated ROS instance.
We describe the individual system components with more detail in the following sections.

\section{Audiovisual Telepresence}

Convincing telepresence requires both seeing and hearing as well as being seen and being heard.
Conventional video conference systems cover this functionality already quite well and it is important
for acceptance that robotic avatar systems do not fall behind these.
We will now detail parts of our system that achieve audiovisual telepresence going beyond
video conferencing while coping with the additional challenges imposed by a robotic avatar system.

\subsection{Robot Cameras \& VR Display}
\label{sec:cameras}
The robot head is equipped with two wide-angle Basler a2A3840-45ucBAS cameras with an optical frame distance of 64\,mm---matching the average human pupillary distance~\citep{dodgson2004variation}.
We crop the stereo video stream to a resolution of 2$\times$2472$\times$2178\,@\,46\,Hz, which gives a
horizontal and vertical FoV of approximately $160^\circ$.
The stream is then compressed
with very low latency on the robot and transmitted over WiFi (see \cref{sec:network}).
The operator wears a Valve Index VR head-mounted display, which renders a view onto the remote scene.
Using a VR HMD leads to full operator immersion.
The cameras are intrinsically and extrinsically calibrated to be able to render the operator view correctly (see \cref{sec:calibration}).

\subsection{6D Movable Head \& Spherical Rendering}

In contrast to all other teams in the ANA Avatar XPRIZE, our robot has a separate 6 Degree of Freedom (DoF) arm
(UFactory xArm6) which carries the head.
This enables the robot to mirror \textit{all} head movements made by the operator, in contrast
to the common pan/tilt neck joints which only allow 2\,DoF rotation.
As a consequence, the operator can look \textit{around} objects simply by translating their head,
as they would if they were present in the remote scene.
They also can choose viewpoints for manipulation that minimize occlusion. Finally, recipients interacting with the avatar frequently note that the full head movements
contribute to the liveliness and identification of the avatar with the operator.

\begin{figure}
 \centering\begin{maybepreview}%
 \begin{tikzpicture}[font=\footnotesize,scale=0.9]
  \begin{scope}
   \begin{scope}
    \draw (2,0) arc (0:180:2) -- cycle;
    \clip (2,0) arc (0:180:2) -- cycle;

    \begin{scope}[rotate=30]
      \draw[fill=green!50]
        (0,0) -- (60:3) arc [radius=3,start angle=60, delta angle=60] -- cycle;
    \end{scope}

    \draw[latex-latex] (15:0.1) -- node[midway,fill=white] {$r$} (15:2);
   \end{scope}
   \draw[fill=white]
     (-0.1,-0.2) rectangle (0.1,0.0)
     (0.0,0.0) -- ++(-0.1,0.1) -- ++(0.2,0.0) -- cycle;

   \begin{scope}[rotate=30]
   \draw[fill=green]
     (-0.1,-0.2) rectangle (0.1,0.0)
     (0.0,0.0) -- ++(-0.1,0.1) -- ++(0.2,0.0) -- cycle;
   \end{scope}
   \node[anchor=north] at (0,-0.2) {(a) Pure rotation};
  \end{scope}
  \begin{scope}[shift={(4.2,0)}]
   \begin{scope}
    \draw (2,0) arc (0:180:2) -- cycle;
    \clip (2,0) arc (0:180:2) -- cycle;

    \begin{scope}[shift={(-0.8,0.8)}, rotate=-30]
       \draw[fill=green!50]
       (0,0) -- (60:4) arc [radius=4,start angle=60, delta angle=60] -- cycle;
    \end{scope}
    \draw[latex-latex] (15:0.1) -- node[midway,fill=white] {$r$} (15:2);
   \end{scope}
   \draw[fill=white]
       (-0.1,-0.2) rectangle (0.1,0.0)
       (0.0,0.0) -- ++(-0.1,0.1) -- ++(0.2,0.0) -- cycle;

   \begin{scope}[shift={(-0.8,0.8)}, rotate=-30]
   \draw[fill=green]
       (-0.1,-0.2) rectangle (0.1,0.0)
       (0.0,0.0) -- ++(-0.1,0.1) -- ++(0.2,0.0) -- cycle;
   \end{scope}

   \draw[-latex] (0,0) to [in=-60,out=170] node [pos=0.9,right,font=\tiny] {$T^C_V$} (-0.8,0.55);

   \node[anchor=north] at (0,-0.2) {(b) Rotation + Translation};
  \end{scope}
 \end{tikzpicture} 
 \end{maybepreview}%
 \caption{%
   Spherical rendering example in 2D. We show only one camera $C$ of the stereo pair,
   the other is processed analogously.
   The robot camera is shown in white with its very wide FoV. The corresponding VR camera $V$,
   which renders the view displayed to the operator, is shown in green.
   The camera image is projected onto the sphere with radius $r$, and then back into the VR camera.
   Pure rotations (a) result in no distortion while translations (b) will
   distort the image if the objects are not exactly at the assumed depth.
   Figure adapted from \citet{schwarz2021vr}.
 }
 \label{fig:spherical}
\end{figure}
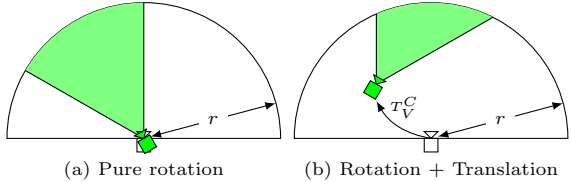

\begin{table}
 \centering
 \caption{Head movement ablations.}\label{tab:head_movement_study}
 \begin{tabular}{lrrr}
  \toprule
  Visual mode & Success & \multicolumn{2}{c}{Completion time [s]} \\
  \cmidrule (lr) {3-4}
              &         & \hspace{2em}Mean & StdDev \\
  \midrule
  \textbb{a) Full 6D} & \textbb{7/7} & \textbb{71.0} & \textbb{50.6} \\
  b) 3D orientation   & 7/7 & 111.7 & 87.5 \\
  c) Fixed perspective & 6/7 & 158.3 & 39.6 \\
  \bottomrule
 \end{tabular}
 Source: \citet{schwarz2021vr}.
\end{table}

\begin{figure*}
	\centering\begin{maybepreview}%
	\begin{tikzpicture}[
	font=\sffamily\tiny,
	inp/.style={draw,minimum width=1cm,minimum height=1cm,inner sep=0},
	inp_nb/.style={draw=none,minimum width=1cm,minimum height=1cm,inner sep=0},
	inp_half/.style={draw=none,minimum width=1cm,minimum height=0.5cm,inner sep=0},
	inp_sm/.style={draw=none,minimum width=0.7cm,minimum height=0.7cm,inner sep=0},
	m/.style={draw,rounded corners,fill=yellow!50,align=center},
	mg/.style={draw,rounded corners,fill=green!40,align=center},
	node distance=0.6,
	every label/.style={font=\sffamily\tiny,align=center,inner sep=2pt,text depth=1pt},
	label distance=0pt,
	]
	
	\node[inp,label=south:Src Img 1] (source_img) {\includegraphics[width=1cm]{images/fig_infer/chris/193/source_0.png}};

	\node[inp,right=0.1 of source_img,label=south:Src KP 1] (source_kp) {\includegraphics[width=1cm]{images/fig_infer/chris/193/kp_source_s4_0.png}};
	
	\node[inner sep=0,left=.3cm of source_img,label={[red,draw,circle,text depth=0,inner sep=1pt]north:A}] (portrait) {
		\includegraphics[width=1cm,clip,trim=20 0 20 0]{images/portrait.png}
	};
	
	\node[m,right=of source_kp] (constr) {Construct\\Driving\\Frame};
	
	\node[inp,right=of constr,label={south:Imaginary\\Driv KP $\hat{I}_D$}] (driving_kp) {\includegraphics[width=1cm]{images/fig_infer/chris/193/kp_driving_s4.png}};
	
	\node[m,above=0.6cm of driving_kp,fill=green!40] (delauny_transform) {Keypoint\\Warp $\mathbf{\psi}$};
	
	\node[inner sep=0,right=2.0cmof driving_kp,label={[yshift=0.6em]south:Motion\\Network $\mathcal{M}$}] (motion) {\tikz[scale=0.7]{
			\draw[fill=yellow!50] (-1,-1) -- (-1,1) -- (0,0.2) -- (1,1) -- (1,-1) -- (0,-0.2) -- cycle;}
	};
	
	\node[inner sep=0,right=1.7cm of motion,label={[yshift=2.5em]south:Generator $\mathcal{G}$}] (generator) {\tikz[scale=0.7]{
			\draw[fill=yellow!50] (-1,-1) -- (-1,1) -- (0,0.2) -- (1,1) -- (1,-1) -- (0,-0.2) -- cycle;}
	};

	\node[inner sep=0pt,anchor=east,label=south:{{\scalefont{0.9} \textcolor{green!50!black}{Warped}}\\[-0.07cm] \textcolor{green!50!black}{{\scalefont{0.9}Mouth Area}}}]
		at ($(delauny_transform-|generator)+(-0.2,0)$) (delauny_mouth) {\includegraphics[width=1cm,clip,trim=0px 0px 0px 150px]{images/fig_infer/chris/193/delauny.png}}; %

	\node[inp,right=1.5 of generator,label=south:Output] (out) {\includegraphics[width=1cm]{images/fig_infer/chris/193/out.png}};

	\node[inp,fill=green!40,font=\normalsize,shape border rotate=90,below=1.8 of generator] (attention) {$\mathcal{A}$};

	\node[inp_half,below=0.6cm of source_img,label] (vr_img) {\includegraphics[width=1cm, clip,trim=1.25cm 0cm 1.25cm 0cm]{images/fig_infer/chris/193/vr.png}};
	\node[inp_half,below=1.33cm of source_img,label=south:VR KP] (vr_kp) {\includegraphics[width=1cm, clip,trim=1.25cm 0cm 1.25cm 0cm]{images/fig_infer/chris/193/kp_vr_s4.png}};
	\coordinate (projin) at ($(vr_kp.east)+(0,-0.2)$);
	\coordinate[m,right=0.5 of projin] (project); %
	
	\node[inner sep=0,left=.3cm of vr_kp,label={[blue,draw,circle,text depth=0,inner sep=1pt]north:C}] (vr_headset) {
		\includegraphics[width=1cm]{images/vr_headset.png}
	};

	\node[m,below=2.7cm of constr] (retrieval) {Image\\Retrieval};
	
	\node[anchor=west,matrix,matrix of nodes,inner sep=0,nodes={inner sep=1pt,minimum width=0.4cm,text depth=-5pt},row sep=0.5pt, column sep=-.5pt]   at ($(vr_kp.west|-retrieval.west)+(0,-0.83)$) (keys) {
		{\includegraphics[width=0.3cm]{images/fig_infer/chris/kp_vr_s4__0.png}} & {\includegraphics[width=0.3cm]{images/fig_infer/chris/kp_vr_s4__24.png}}  &
		{$\dots$}& {\includegraphics[width=0.3cm]{images/fig_infer/chris/kp_vr_s4__48.png}} &
		{\includegraphics[width=0.3cm]{images/fig_infer/chris/kp_vr_s4__72.png}} \\
		{\includegraphics[width=0.3cm]{images/fig_infer/chris/kp_vr_s4__96.png}} & {\includegraphics[width=0.3cm]{images/fig_infer/chris/kp_vr_s4__120.png}}  &
		{$\dots$}& {\includegraphics[width=0.3cm]{images/fig_infer/chris/kp_vr_s4__144.png}} & {\includegraphics[width=0.3cm]{images/fig_infer/chris/kp_vr_s4__298.png}} \\
	};
	\draw[ decorate, decoration = {calligraphic brace}, ultra thick] ($(keys.north east)+(0.1,0)$) -- ($(keys.south east)+(0.1,0)$);
	
	\node[anchor=west,matrix,matrix of nodes,inner sep=0,nodes={inner sep=1pt,minimum width=0.4cm,text depth=-5pt},row sep=0.5pt, column sep=-.5pt] at (vr_kp.west|-retrieval.west) (values) {
		{\includegraphics[width=0.3cm]{images/fig_infer/chris/face_0.png}} & {\includegraphics[width=0.3cm]{images/fig_infer/chris/face_24.png}} &
		{$\dots$}& {\includegraphics[width=0.3cm]{images/fig_infer/chris/face_48.png}} & {\includegraphics[width=0.3cm]{images/fig_infer/chris/face_72.png}} \\
		{\includegraphics[width=0.3cm]{images/fig_infer/chris/face_96.png}} & {\includegraphics[width=0.3cm]{images/fig_infer/chris/face_120.png}}  &
		{$\dots$}& {\includegraphics[width=0.3cm]{images/fig_infer/chris/face_144.png}} & {\includegraphics[width=0.3cm]{images/fig_infer/chris/face_298.png}} \\
	};
	\draw[decorate, decoration = {calligraphic brace},ultra thick] ($(values.north east)+(0.1,0)$) -- ($(values.south east)+(0.1,0)$);
	
	\coordinate (storage) at ($(keys.west)!0.5!(values.west)$);
	\node[inner sep=0,left=.3cm of storage,label={[green!50!black,draw,circle,text depth=0,inner sep=1pt]north:B}] (portrait2) {
		\includegraphics[width=1cm,clip,trim=20 0 20 0]{images/portrait.png}
	};
	
	\node[inp] at ($(retrieval.east)+(1.15,-0.2)$) (retrieved_kp) {\includegraphics[width=1cm]{images/fig_infer/chris/193/kp_source_s4_4.png}};
	\node[inp,right=0.1 of retrieved_kp] (retrieved_img){\includegraphics[width=1cm]{images/fig_infer/chris/193/source_4.png}};
	\node[every label,anchor=north] at ($(retrieved_kp.south)!0.5!(retrieved_img.south)$) {Expressional Source \\ KPs+Imgs};
	
	\node[inp] at ($(retrieved_img)+(0,1.48)$) (source_img2){\includegraphics[width=1cm]{images/fig_infer/chris/193/source_2.png}};
	\node[] at ($(retrieved_img)+(0,0.85)$) (source_dots){$\vdots$};
	
	\node[inp] at ($(retrieved_kp)+(0,1.48)$) (source_kps2){\includegraphics[width=1cm]{images/fig_infer/chris/193/kp_source_s4_2.png}};
	\node[] at ($(retrieved_kp)+(0,0.85)$) (source_dots2){$\vdots$};
	
	\draw[black,decorate,decoration={calligraphic brace,amplitude=6pt}, thick] ($(source_kps2.north west)+(-0.5pt,1pt)$)--($(source_img2.north east)+(0.5pt,1pt)$) node[midway, above](brace){};
	\node[draw,inner sep=1pt,anchor=north east,label=south:Avatar Display] (display) at ($(out.south east)+(-0.1,-1.15)$)
	{\includegraphics[width=2.2cm,clip,trim=0 240 50 160]{images/fig_infer/avatar_display2.jpg}};
	
	\draw[black,decorate,decoration={calligraphic brace,amplitude=6pt}, thick] ($(source_img2.north east)+(1pt,0pt)$) -- ($(retrieved_img.south east)+(1pt,0pt)$) node[midway, above](brace2){};
	
	\draw[pen colour={black!50!green},decorate,decorate,decoration={calligraphic brace,amplitude=6pt},ultra thick] ($(keys.south east)+(0pt,-0.1pt)$) -- ($(keys.south west)+(0pt,-0.1pt)$) node[midway, above](brace_kp){};

	\begin{scope}[-latex]
	\draw (source_img.north) -- ($(source_img.north|-delauny_transform.south)+(0,-0.2)$) -| ($(generator.west)+(-1.2,0.4)$)--($(generator.west)+(0,0.4)$);
	\draw (source_kp) -- (constr);
	\draw (source_kp.east) -- ++(0.3,0) -- ++(0,0.6) -| ($(driving_kp.east)!0.5!(motion.west)+(0,0.3)$) -- ($(motion.west)+(0,0.3)$);
	\draw (constr) -- (driving_kp);
	\draw (driving_kp) -- (motion);
	\draw[line width=0.25mm, black!50!green] (driving_kp) -- (delauny_transform);
	\draw[line width=0.25mm, black!50!green] (delauny_transform) -- (delauny_mouth);
	\draw[line width=0.25mm, black!50!green] (delauny_mouth) -| ($(generator.north)+(0,-0.4)$);
	\draw[line width=0.25mm, black!50!green] ($(vr_img.west)+(-0.05,0.15)$) --++(-1.3,0) |- (delauny_transform.west);
	\draw (motion.east) -- (generator);
	\draw (generator) -- (out);
	\draw[magenta, line width=0.35mm] (out.south) ++(0,-0.35) |- (rel2 cs:x=63,y=65,name=display);

	\draw[line width=0.35mm, black!50!green] (attention) -- (generator.south)--++(0,0.4) node [black,pos=-1.4,left,inner sep=1pt,black!50!green] {${{a_S}_i}$}; 

	\draw (vr_kp.east|-project) -| (constr);
	\draw ($(vr_kp.east)+(0,0.2)$) -| ($(constr.south)+(-0.2,0)$) node [pos=0.25,above,inner sep=1pt] {Eye Coord.};
	\draw (project) -| (constr) node [pos=0.215,above,inner sep=1pt]{Lower Face KP};
	\draw [dashed](project -| constr) -- (retrieval) node [pos=0.75,right] {Query};
	\draw [dashed] (values.east) ++(0.25,0) -- (retrieval) node [midway,above] {V};
	\draw [dashed] (keys.east) ++(0.25,0) -| (retrieval) node [pos=0.75,right] {K};
	\draw [dashed] (retrieval) -- ($(retrieved_kp.west)+(0,0.2)$);
	\draw (brace.north) -- ++(0,0.3) -| ($(driving_kp.east)!0.5!(motion.west)+(0,-0.3)$) -- ($(motion.west)+(0,-0.3)$);
	\draw ($(brace2.east)+(0.15,-0.125)$) --++(1,0) --++(0,1) --++ (1.5,0) |- ($(generator.west)+(0,-0.4)$);
	\draw [line width=0.35mm, black!50!green] ($(brace_kp.south)+(0,-0.2)$) -- ($(brace_kp.south)+(0,-0.3)$) -| (attention.south) node [black,left,inner sep=1pt,black!50!green,pos=0.65] {$K_{\mathcal{VR}}(I_{S_i})$};
	\draw [line width=0.35mm, black!50!green] ($(attention.west)+(-1.6,0)$) -- (attention) node [black,pos=0.4,above,inner sep=1pt,black!50!green] {$K_{\mathcal{VR}}(\hat{I}_{D})$};
	\end{scope}

	\begin{pgfonlayer}{background}
	\draw[fill=blue!18,rounded corners,draw=none]
	($(source_img.north west)+(-0.2,0.2)$) -|
	($(source_kp.south east)+(0.2,-0.35)$) -|
	($(source_img.south west)+(-0.2,-0.35)$)-- cycle;
	\end{pgfonlayer}
	
	\begin{pgfonlayer}{background}
	\draw[fill=green!30,rounded corners,draw=none]
	($(source_kps2.north west)+(-0.25,0.25)$) -|
	($(retrieved_img.south east)+(0.25,-0.65)$) -|
	($(retrieved_kp.south west)+(-0.25,-0.65)$)-- cycle;
	\end{pgfonlayer}
	\end{tikzpicture} 
     \end{maybepreview}%
	\caption{Inference pipeline for VR Facial Animation. New components compared to semifinals~\cite{baseline} are highlighted in green. We select 4-5 still source images from a portrait video of the operator shot before the run as source images~\textcolor{red}{(A)}. The remaining frames are optionally used as a key-value storage of retrievable expression keypoints and
		corresponding images~\textcolor{green!50!black}{(B)}.
		The live keypoints measured inside and outside the VR headset~\textcolor{blue}{(C)} are then projected to the first
		source image frame, where they are optionally used to retrieve the closest expression image with keypoints from the storage.
		The keypoints of all source images, including the retrieved one, and a constructed set of driving keypoints then
		enter the motion network $\mathcal{M}$, which estimates a warping grid that is used to deform the source images features, extracted by the generator-encoder network, to match the driving keypoints.
		The deformed features are aggregated over the source images in the lower facial area using a trainable attention mechanism $\mathcal{A}$.
		The mouth camera image from the HMD is warped into the lower facial area of the constructed driving keypoints and then encoded by a separate encoder network. An estimated mask gates the aggregated deformed source features using the warped mouth camera features.
		The masked aggregated features are then decoded to produce the output.
	}
	\label{fig:facial_animation_pipeline}
\end{figure*}

Allowing head movements, comes with a problem, though. Since there is considerable movement latency
(roughly 200\,ms) introduced by the masses, friction, and motor velocity constraints, directly rendering
the video stream on the HMD results in operator motion sickness.
Instead, we use a technique which renders each frame in a sphere centered on the capture location (see \cref{fig:spherical}).
The operator moves freely and with low latency inside this sphere.
While head translations induce transient distortions (see \cref{fig:spherical}), head rotations, which are most affected
by latency due to the large lever effect for distant objects, can be perfectly handled.
For more details about this technique, we refer to \citet{schwarz2021vr}.

We evaluated the 6D neck joint against ablations with only 3D rotation or a fixed perspective in
a small user study, which showed advantages for the full 6D mode (see \cref{tab:head_movement_study}).
Additionally, operators reported that the freedom of movement in 6D was helpful for an insertion task,
where viewing from the side was beneficial~\citep{schwarz2021vr}.
The benefits of stereoscopic vision itself for teleoperation have been shown before, e.g.
by \citet{triantafyllidis2020study}.

\subsection{Facial Animation}
\label{sec:facial_animation}

\begin{figure}
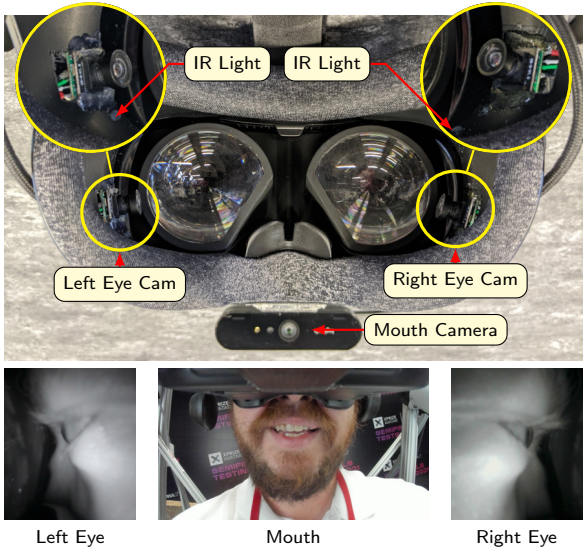

\begin{maybepreview}%
\begin{tikzpicture}[
	font=\footnotesize\sffamily,
	box/.style={m,draw=black,thin},
	lab/.style={draw=red,latex-,thick}
]

\begin{pgfonlayer}{background}
\node[anchor=south west,inner sep=0] (img) at (0,0) {\includegraphics[width=\linewidth,angle=180,clip,trim=0 400 0 200]{images/vr.jpg}};
\end{pgfonlayer}
\draw[lab] (rel cs:x=.53,y=.09,name=img) -- ++(0.7,0) node [box,anchor=west] {Mouth Camera};

\node[circle,minimum width=1cm,draw=yellow,very thick] (right_eye) at (rel cs:x=0.78,y=.44,name=img) {};
\draw[lab] (right_eye.south) -- ++(0,-0.2) node [box,anchor=north] {Right Eye Cam};
\begin{scope}[shift={(rel cs:x=.85,y=.82,name=img)}]
  \clip (0,0) circle (1.0cm);
  \node[rotate=220] at (0.2,0.29) {\includegraphics[width=3.0cm]{images/right_eye_cam.jpg}};
  \node[circle,minimum width=1.96cm,draw=yellow,very thick] (right_eye_circle) at (0,0) {};
\end{scope}
\draw[yellow,thick] (right_eye) -- (right_eye_circle);

\node[circle,minimum width=1cm,draw=yellow,very thick] (left_eye) at (rel cs:x=.20,y=.43,name=img) {};

\draw[lab] (left_eye.south) -- ++(0,-0.2) node [box,anchor=north] {Left Eye Cam};
\begin{scope}[shift={(rel cs:x=.15,y=.82,name=img)}]
  \clip (0,0) circle (1.0cm);
  \node[rotate=160] at (0,0) {\includegraphics[width=3.0cm]{images/left_eye_cam.jpg}};
  \node[circle,minimum width=1.96cm,draw=yellow,very thick] (left_eye_circle) at (0,0) {};
\end{scope}

\draw[yellow,thick] (left_eye) -- (left_eye_circle);

\draw[lab] (rel cs:x=.192,y=.705,name=img) -- ++(45:1cm) -- ++(0.2,0) node (ir1lab) [box,anchor=west] {IR Light};

\coordinate (ir2) at (rel cs:x=.775,y=.667,name=img);
\draw[lab] let \p1=(ir2), \p2=(ir1lab) in (ir2) -- ++(\y1-\y2,\y2-\y1) -- ++(-0.2,0) node [box,anchor=east,draw=black,thin] {IR Light};

\node[anchor=north west,inner sep=0,label=south:Left Eye] at ($(img.south west)+(0,-0.1)$) {\includegraphics[height=2cm,clip,trim=0 100 0 100]{images/vr_cams/left.png}};
\node[anchor=north east,inner sep=0,label=south:Right Eye] at ($(img.south east)+(0,-0.1)$) {\includegraphics[height=2cm,clip,trim=0 100 0 100]{images/vr_cams/right.png}};
\node[anchor=north,inner sep=0,label=south:Mouth] at ($(img.south)+(0,-0.1)$) {\includegraphics[height=2cm]{images/vr_cams/mouth.png}};
\end{tikzpicture} 
 \end{maybepreview}%
\caption{Modified Valve Index VR headset. We attached three additional cameras to capture the eyes and the mouth expression of the operator. We show the corresponding camera views at the bottom. Source: \citet{baseline}.}
\label{fig:hmd}
\end{figure}

Making the operator seen properly is challenging, since they are wearing a VR headset.
Simply capturing and displaying a video stream would be possible, but falls behind video conferencing solutions
as facial expressions are partially hidden and distorted.
Instead, we reconstruct the operator's face from video data captured by a mouth camera and two eye cameras mounted on the HMD (see \cref{fig:hmd}).
Unlike related methods~\citep{lombardi2018deep,vr_facial,faceaudio}, we do not require per-operator training for an unseen person.
Instead, our method is trained on large speaking-head datasets and generalizes to new operators.

\Cref{fig:facial_animation_pipeline} gives an overview of our pipeline.
We first capture a source video of an operator without the HMD and another one from the mouth camera on the HMD. From the first source video, we select four different fixed source images and optimize a keypoint mapping which projects keypoints from the mouth camera into the first source image, taking into account facial deformations caused by the HMD. Given the projected keypoints and eye tracking results, we dynamically construct imaginary driving keypoints.

The encoded features of the source images and a dynamically retrieved expression image are deformed into the constructed driving keypoints using a deformation grid predicted by the motion network.
We prevent temporal inconsistencies, which are mainly caused by expression image changes, with a source image attention mechanism and visual mouth camera guidance. Both are applied to the latent space before decoding the features to produce the output image.
We will give an overview of these two innovations below. For a detailed explanation we refer to~\citet{rochow2023attention}.

\paragraph{Source Image Attention Mechanism}
We select several different source images and train a source image attention mechanism that equips the network with the ability to decide how much information it requires from each source image. The attention values are then used to aggregate the latent representations of the source images after deforming the features to align them in the constructed driving keypoints.
This significantly improves temporal consistency compared to our semifinal solution~\cite{baseline}, as the attention values are estimated by a continuous function that smoothly adapts to changes in the mouth camera stream. Our previous approach, however, only utilized one retrieved expression image that can change abruptly and therefore introduce strong discontinuous effects.
More diverse facial expressions of an operator are presented to the network which reduces the network's dependence to the retrieved expression image and improves animation quality.

\paragraph{Visual Mouth Camera Guidance}
Including direct visual information from the mouth camera allows to resolve keypoint ambiguities and contains a broader range of facial expressions.
However, it is difficult to train this part since large-scale datasets with mouth camera \& full face images are not available
(consider that the VR headset also visibly deforms the face, so a simple crop does not suffice).

We enable visual mouth camera guidance by estimating a Delaunay triangulation in the lower facial keypoints and using the barycentric coordinates to sample the mouth camera image in the lower facial area of the driving keypoints. This roughly aligns both representations. A trainable encoder network then estimates a latent representation and conditions the aggregated deformed source image features via gated convolutions~\citep{gated_convs}.
The masking operation in a gated convolution allows the elimination of poor image features that do not correspond to visual mouth camera information, while still encoding additional information in the latent representation.  
Another important advantage is that direct information propagation is prevented, which would lead to the network ``pasting'' the mouth section without correction. This allows us to continue training with entire faces.
During training, we utilize the lower facial area of the driving image as the mouth camera input and simulate the imperfect perspective transformation by adding keypoint noise in both the source and target keypoints. In addition, different types of image noise are added to account for different lighting conditions. This, combined with the source image attention mechanism, already improves performance compared to our semifinal solution (see Semi vs. Ours-NF in \cref{tab:eval:ablations}).
However, even when we emulate the effect of directly transforming the mouth camera keypoints to the driving keypoints, there are still differences that limit performance.
We address this by manually searching for correspondences between mouth camera images and entire faces to annotate a small, suitable VR dataset consisting of 13 different persons. During finetuning, we select samples from this annotated set with a probability of 6\%.
This significantly improves performance.

\begin{figure}
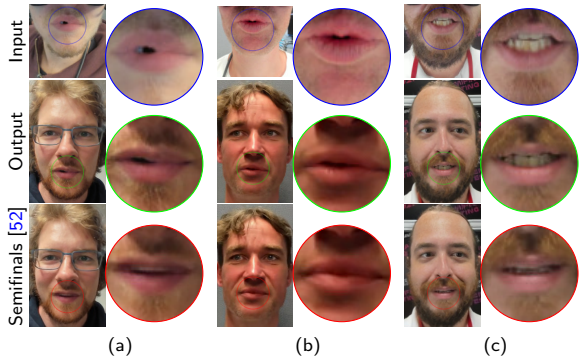

	\centering\begin{maybepreview}%
	\begin{tikzpicture}[font=\sffamily\footnotesize, spy using outlines={circle, magnification=3.0, size=0.08\textwidth,every spy in node/.append style={thin,opacity=1}}]
	\node[inner sep=0,matrix, matrix of nodes, nodes={inner sep=0,anchor=center}, row sep=1pt, column sep=42pt,
	column 1/.style={anchor=center, column sep=1pt},
	column 2/.style={anchor= center},
	column 3/.style={anchor= center},
	column 4/.style={anchor=center}] (m) {
		|[rotate=90]| Input &
		|(in1)|\includegraphics[width=0.062\textwidth,clip,trim=9cm 0cm 9cm 0cm]{images/vgl_infer/andre_2_chris/in/vr_238.png} &    
		|(in3)|\includegraphics[width=0.062\textwidth,clip,trim=13cm 0cm 13cm 0cm]{images/vgl_infer/mischa/in/vr_48.png} &
		|(in5)|\includegraphics[width=0.062\textwidth,clip,trim=9cm 0cm 9cm 0cm]{images/vgl_infer/judge2/in/vr_303.png}\\
		|[rotate=90]| Output &
		|(out1)|\includegraphics[width=0.062\textwidth,clip,trim=1.75cm 0cm 1.75cm 0cm]{images/vgl_infer/andre_2_chris/ours25/out_238_eye.png} &
		|(out2)|\includegraphics[width=0.062\textwidth,clip,trim=1.75cm 0cm 1.75cm 0cm]{images/vgl_infer/mischa/ours25/out_48_eye.png} &
		|(out3)|\includegraphics[width=0.062\textwidth,clip,trim=1.75cm 0cm 1.75cm 0cm]{images/vgl_infer/judge2/ours25/out_303_eye.png}\\
		|[rotate=90]| Semifinals~\citep{baseline} &
		|(out1_b)|\includegraphics[width=0.062\textwidth,clip,trim=1.75cm 0cm 1.75cm 0cm]{images/vgl_infer/andre_2_chris/base+TCF/out_base_238_eye.png} &
		|(out2_b)|\includegraphics[width=0.062\textwidth,clip,trim=1.75cm 0cm 1.75cm 0cm]{images/vgl_infer/mischa/base+TCF/out_48_eye.png} &
		|(out3_b)|\includegraphics[width=0.062\textwidth,clip,trim=1.75cm 0cm 1.75cm 0cm]{images/vgl_infer/judge2/base+TCF/out_303_eye.png} \\
	};
	\coordinate (a) at ($(in1.north)!0.5!(in3.north)$);

	\spy [red, opacity=0.35,every spy on node/.append style={ultra thin}] on (rel2 cs:x=50,y=25,name=out1_b) in node [right] at ($(out1_b.east)+(0,-0.1)$);
	\spy [blue,size=0.08\textwidth, magnification=2.6, opacity=0.3,every spy on node/.append style={ultra thin}] on (rel2 cs:x=50,y=71,name=in1) in node [right] at ($(out1.north east)+(0,0.3)$);
	\spy [green, opacity=0.3,every spy on node/.append style={ultra thin}] on (rel2 cs:x=50,y=25,name=out1) in node [right] at ($(out1.east)+(0,-0.3)$);
	
	\spy [red, opacity=0.35,every spy on node/.append style={ultra thin}] on (rel2 cs:x=47.5,y=28,name=out2_b) in node [right] at ($(out2_b.east)+(0,-0.1)$);
	\spy [blue,size=0.08\textwidth, magnification=2.3, opacity=0.3,every spy on node/.append style={ultra thin}] on (rel2 cs:x=47,y=66,name=in3) in node [right] at ($(out2.north east)+(0,0.3)$);
	\spy [green, opacity=0.3,every spy on node/.append style={ultra thin}] on (rel2 cs:x=47.5,y=28,name=out2) in node [right] at ($(out2.east)+(0,-0.3)$);
	
	\spy [red, opacity=0.3,every spy on node/.append style={ultra thin}] on (rel2 cs:x=51.5,y=28,name=out3_b) in node [right] at ($(out3_b.east)+(0,-0.1)$);
	\spy [blue,size=0.08\textwidth, magnification=2.3, opacity=0.35,every spy on node/.append style={ultra thin}] on (rel2 cs:x=50,y=67,name=in5) in node [right] at ($(out3.north east)+(0,0.3)$);
	\spy [green, opacity=0.3,every spy on node/.append style={ultra thin}] on (rel2 cs:x=51.5,y=28,name=out3) in node [right] at ($(out3.east)+(0,-0.3)$);
	
	\node[anchor=north] at ($(out1_b.south west)+(1.2,0)$) {(a)};
	\node[anchor=north] at ($(out2_b.south west)+(1.2,0)$) {(b)};
	\node[anchor=north] at ($(out3_b.south west)+(1.2,0)$) {(c)};

	\end{tikzpicture} 
     \end{maybepreview}%
	\caption{%
	Qualitative facial animation results. We show mouth camera input, the animated face, and results obtained by our semifinal solution~\cite{baseline}
	for three cases:
	(a) Transfer from a different operator in a particularly challenging case with sticking lips,
	(b) slightly open mouth, and
	(c) teeth showing.
	}
	\label{fig:q_fancy}
\end{figure}

\paragraph{Facial Animation Evaluation}

\begin{table}
	\centering \footnotesize
	\caption{Facial animation ablations}
	\label{tab:eval:ablations}
	\begin{tabular}{l@{\hspace{12pt}}ccc@{\hspace{7pt}}@{\hspace{7pt}}r} \toprule
		Method       & PSNR$\uparrow$    & SSIM$\uparrow$     &LPIPS$\downarrow$   & TIC$\downarrow$ \\ \midrule
		Ours         & \textbb{28.75}   & \textbb{0.8586}    & \textbb{0.0361}  & +8.1\%    \\
		Ours-5-Fix   & 28.50   & 0.8504    & 0.0376  & +\textbb{0.0}\%  \\
		Ours-NF      & 27.20   & 0.8369    & 0.0465  & +3.7\%  \\  %
		Semi~\citep{baseline}& 25.10   & 0.7809    & 0.0580  & +98.9\%  \\

		\bottomrule
	\end{tabular} \vspace{1.5ex}
	\textit{Ours-5-Fix}: only five fixed source images without image retrieval.
	\textit{Ours-NF}: No finetuning on manually annotated mouth camera images.
	\textit{Semi}: Our semifinal solution. TIC: temporal inconsistency measure normalized to Ours-5-FIX. For details, we refer to \citep{rochow2023attention}.
\end{table}

We evaluate our method variations and compare them to our previous method~\cite{baseline} on an annotated dataset of five unseen individuals, in which we manually assign HMD mouth camera images to roughly corresponding facial images. 
Accuracy and temporal inconsistency are reported in \cref{tab:eval:ablations}. The results indicate that all our method ablations outperform our previous work in terms of temporal consistency and accuracy. 
The highest accuracy is achieved when the last expression image is dynamically retrieved from the source video during inference (Ours).
Temporal consistency, however, is maximized when all five source images are fixed during inference (Ours-5-Fix).  
This highlights the trade-off between accuracy and temporal consistency.
We also show qualitative examples in \cref{fig:q_fancy}.

\subsection{Audio}
\label{sec:audio}

\begin{figure}[b]
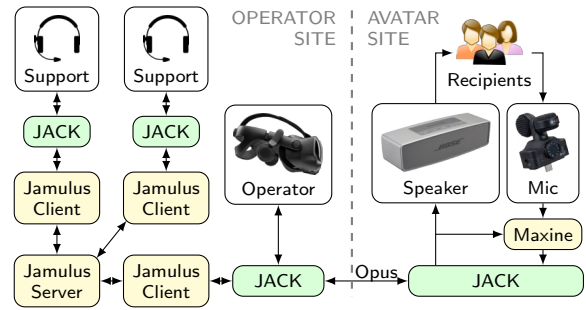

 \centering\begin{maybepreview}%
 \begin{tikzpicture}[every node/.append style={text depth=0pt}]
  \node[m,fill=green!15!white,minimum width=1.2cm] (opjack) {JACK};
  \node[m,above=0.8 of opjack,fill=white] (hmd) {\includegraphics[height=0.8cm]{images/audio_system/vr_headset.png}\\Operator};

  \node[m, left=0.3 of opjack.west] (opjam) {Jamulus\\Client};
  \node[m, anchor=south east, left=0.3 of opjam] (jam) {Jamulus\\Server};
  \node[m, above=0.4 of jam] (crew1) {Jamulus\\Client};
  \node[m, above=0.4 of opjam] (crew2) {Jamulus\\Client};
  \node[m,fill=green!15!white,above=0.32 of crew1] (crew1_jack) {JACK};
  \node[m,fill=green!15!white,above=0.32 of crew2] (crew2_jack) {JACK};

  \node[m,above=0.32 of crew1_jack,fill=white] (head1) {\includegraphics[height=0.6cm]{images/headset.png}\\Support};
  \node[m,above=0.32 of crew2_jack,fill=white] (head2) {\includegraphics[height=0.6cm]{images/headset.png}\\Support};

  \node[m,fill=green!15!white,right=1.1 of opjack,minimum width=2.3cm] (ajack) {JACK};
  \node[m,anchor=south west] (maxine) at ($(ajack.north)+(+0.1,0.2)$) {Maxine};
  \node[m,above=0.2 of maxine,fill=white] (mic) {\includegraphics[height=0.8cm]{images/audio_system/mic.png}\\Mic};
  \coordinate (ajackp) at ($(ajack.north)+(0.05,0)$);
  \node[m,anchor=east,fill=white] (speaker) at (ajackp|-mic) {\includegraphics[height=0.8cm]{images/audio_system/speaker.png}\\Speaker};
  
  \coordinate (mid_speaker_mic) at ($(mic.north)!0.5!(speaker.north)$);
  \node[align=center,anchor=south] (recp) at ($(mid_speaker_mic)+(0,0.1)$) {\includegraphics[height=0.6cm]{images/people.png}\\Recipients};

  \begin{scope}[latex-latex]
    \draw (opjack) -- (hmd);
    \draw (ajack) -- (opjack) node[pos=0.38,above,inner sep=1pt] {Opus};
    \draw (opjack) -- (opjam);
    \draw (opjam) -- (jam);

    \draw (crew1_jack) -- (head1);
    \draw (crew2_jack) -- (head2);
    \draw (crew1) -- (crew1_jack);
    \draw (crew2) -- (crew2_jack);
    \draw (crew1.south) -- (jam.north);
    \draw ($(crew2.south west)+(0.05,0.05)$) -- ($(jam.north east)-(0.05,0.05)$);;

    \draw[latex-] (ajack.north-|maxine) -- (maxine);
    \draw[-latex] (mic) -- (maxine);
    \draw[latex-] (speaker) -- (speaker|-ajack.north);
    \draw[-latex] (speaker|-maxine) |- (maxine);

    \draw[-latex] (speaker.north) |- ($(recp.west)+(0.2,0)$);
    \draw[latex-] (mic.north) |- ($(recp.east)-(0.12,0)$);
  \end{scope}

  \coordinate (mid) at ($(hmd.east)!0.5!(speaker.west)$);
  \coordinate (midu) at (mid|-recp.north);
  \coordinate (midl) at (mid|-opjack.south);
  \begin{scope}[draw=black!50,text=black!50]
    \node[anchor=north west,align=left,inner sep=0pt] at ($(midu)+(0.2,0)$) {AVATAR\\SITE};
    \node[anchor=north east,align=right,inner sep=0pt] at ($(midu)+(-0.2,0)$) {OPERATOR\\SITE};
    \draw[dashed,thick] (midu) -- (midl);
  \end{scope}
 \end{tikzpicture}
 \end{maybepreview}%
 \caption{Audio system. The setup allows multiple support crew members to listen in and to
 communicate with both the operator and recipient(s).}
 \label{fig:audio_system}
\end{figure}

Auditive perception and communication capabilities are key modalities for an immersive telepresence experience.
A central objective for the design of the audio system is to optimize for low latency while maintaining the integrity of high-resolution stereo audio.
\Cref{fig:audio_system} shows an overview of our audio solution.
The audio hardware on the operator station is comprised of the built-in microphone and headphones of the HMD, 
while the avatar robot is equipped with a stereo microphone mounted on top of its head display and a loudspeaker attached to its torso (see \cref{fig:sys_overview}).
The directionality of both devices on the avatar side, combined with the rigid connection between the microphone and the 6D movable head,
establish a human-like experience w.r.t. room acoustics for both the operator and the recipients. 
All audio devices are connected utilizing the JACK audio connection kit\footnote{\url{https://jackaudio.org}}, matching the requirements for high resolution (24\,bit\,\&\,48\,kHz) and low latency (512 samples),
while also providing the flexibility to establish robustness through layers of self-recovery, monitoring, and control.

The most expensive step in terms of latency is the WiFi transmission to and from the avatar robot.
Therefore, we encode all audio packages using the OPUS audio codec and transmit them via UDP redundantly over the 2.4\,GHz and 5.0\,GHz WiFi networks (see \cref{sec:network}).
Depending on the concrete network latency and the set volume of the avatar's loudspeaker, the operator can be exposed to hearing an echo of their voice feeding back from the loudspeaker to the microphone.
To combat this, we integrate an echo cancellation system based on NVIDIA Maxine\footnote{\url{https://developer.nvidia.com/maxine}}, which also allows reduction of noise and reverberation.
Finally, we integrate Jamulus\footnote{\url{https://jamulus.io}} to provide audio conferencing functionality, augmenting the communication between the operator station and the avatar robot.
In particular, each team member can wear headphones and join the audio conference to communicate with each other, with the operator, and with recipients through the avatar.
This makes it easy for everyone involved to communicate and keep track of the current status,
which is particularly helpful during setup, training, and monitoring.

\section{Telemanipulation}
\label{sec:telemanipulation}

\begin{figure}
\centering\begin{maybepreview}%
 \includegraphics[height=3.1cm, clip, trim=0px 100px 0px 100px]{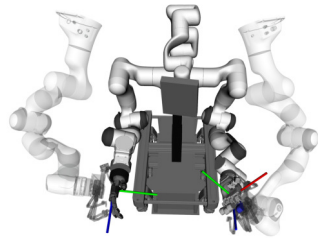}
 \end{maybepreview}%
 \caption{Kinematic arm configuration for both the avatar robot (solid model) and the operator station (transparent model). The axes represent the common hand frame.}
 \label{fig:arm_configuration}
\end{figure}

Telemanipulation allows the operator to interact with remote environments and is a key element of our avatar system.
The operator station and the avatar robot each use two Franka Emika 7\,DoF Panda arms for this purpose.
The operator wears SenseGlove DK1 hand exoskeletons which are mounted via Nordbo NRS-6050-D80 force-torque sensors to the Panda arms (see \cref{fig:sys_overview}).
This allows to track operator finger and hand positions and provides force and haptic feedback (see \cref{sec:arm_controller,sec:hand_controller}).

\begin{figure}
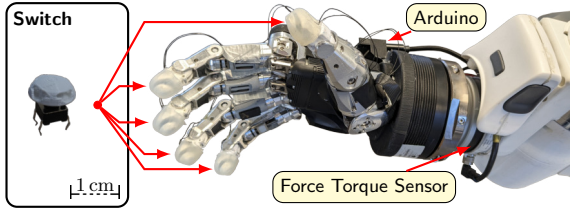

\centering\begin{maybepreview}%
\begin{tikzpicture}[l/.style={anchor=south west, font=\sffamily\scriptsize, scale=\textscale}, b/.style={anchor=south west, draw, rounded corners, align=center, fill=white, font=\sffamily\scriptsize, scale=\textscale}]
	  \definecolor{background1}{rgb}    {0.0,0.0,0.0}
	 
	  \def\boxhgap{0.1};
	  \def\boxheight{2.5};
	  \def\boxalen{1.6};
	  \def\textscale{0.8};

	  \node[inner sep=0pt] (img1) {\includegraphics[width=.75\linewidth]{images/telemanipulation/svh.png}};
	  
	  \draw[thick, background1, rounded corners] ($(img1.south west)+(-\boxhgap-\boxalen,-0.2cm)$) -- ($(img1.south west)+(-\boxhgap-\boxalen,\boxheight)$) -- ($(img1.south west)+(-\boxhgap,\boxheight)$) -- ($(img1.south west)+(-\boxhgap,-0.2cm)$) -- cycle;
	  \node[anchor = north west, font=\sffamily\scriptsize, scale=\textscale] at ($(img1.south west)+(-\boxhgap-\boxalen,\boxheight)-(0.01,0.01)$) {\textbf{Switch}};
	  \node(scale)[inner sep=0pt, anchor=south east] at ($(img1.south west)+(-\boxhgap,-0.2cm)+(-0.1,0.1)$) {\includegraphics[scale=0.08]{images/roughness_hardware/hardware_10mm.png}};
	  \node[font=\sffamily\scriptsize, scale=\textscale, above left=-0.18cm and -0.68cm of scale] {$\SI{1}{\cm}$};
	  \node(switch)[inner sep=0pt, anchor=center] at ($(img1.west)+(-1.1,-0.1)$) {\includegraphics[scale=0.16]{images/telemanipulation/switch.png}};
	  
	  \node(origin)[inner sep=0pt, draw=red, fill=red, circle, minimum size=.1cm] at ($(img1.west)+(-0.5,-0.1)$) {};

	  \draw[-latex,thick,color=red] (origin) -- ($(img1.north east)-(5.60,0.25)$) -- ($(img1.north east)-(3.75,0.25)$);
	  \draw[-latex,thick,color=red] (origin) -- ($(img1.north east)-(5.93,1.10)$) -- ($(img1.north east)-(5.57,1.10)$);
	  \draw[-latex,thick,color=red] (origin) -- ($(img1.north east)-(5.93,1.58)$) -- ($(img1.north east)-(5.54,1.58)$);
	  \draw[-latex,thick,color=red] (origin) -- ($(img1.north east)-(5.60,2.00)$) -- ($(img1.north east)-(5.25,2.00)$);
	  \draw[-latex,thick,color=red] (origin) -- ($(img1.north east)-(5.60,2.20)$) -- ($(img1.north east)-(4.70,2.20)$);

	  \node(label_a)[font=\sffamily\scriptsize, scale=\textscale, anchor=north east, below left=0.0cm and 1.1cm of img1.north east, draw=black, rounded corners, m, thin] {Arduino};
	  \node(label_b)[font=\sffamily\scriptsize, scale=\textscale, anchor=south east, above left=-0.15cm and 1.5cm of img1.south east, draw=black, rounded corners, m, thin] {Force Torque Sensor}; %
	  
	  \draw[-latex,thick,color=red] (label_a) -- ($(img1.north east)-(2.40,0.5)$);
	  \draw[-latex,thick,color=red] (label_b) -- ($(img1.north east)-(1.30,1.9)$);

	\end{tikzpicture} 
  \end{maybepreview}%
 \caption{Right Schunk SVH hand with custom fingertips holding pushbutton switches for contact measurements.}
 \label{fig:svh}
\end{figure}

The arms of the avatar robot are mounted in an anthropomorphic configuration to match the human arm workspace as closely as possible while 
minimizing the shoulder width to 78\,cm, allowing for easy navigation through narrow passages (see \cref{fig:arm_configuration}).
Since the Panda arms are neither symmetrical nor available in a mirrored version, the right hand is mounted in a 90\,$^{\circ}$ angle to avoid reaching the limit of joint five during normal manipulation tasks.
OnRobot HEX-E force-torque sensors are mounted on the arms that hold a Schunk SVH and SIH hand on the right and left sides, respectively (see \cref{fig:svh,fig:vr_view}).
Having two different hands on the avatar robot increases the manipulation capabilities for the operator: The active 9\,DoF SVH hand allows very dexterous
manipulation but has a rather low payload of about 1.5\,kg.
In contrast, the cable-driven 5\,DoF Schunk SIH is advantageous for less precise but more forceful tasks.
We modified all four Panda arms both in hard- and software to support our requirements:
The firmware was customized to allow non-horizontal mounting.
In addition, the modified firmware allows to automatically recover from error states under supervision of the control PC.
This is one crucial feature that greatly increased our system robustness (see \cref{sec:sysmon}).
In addition, we decreased the size of the avatar wrists by removing unused buttons, 3D printing smaller covers for the last joint, and mounting the teach buttons to a different location.
This reduces collisions, e.g. when manipulating objects on a table.

More details about the arm and hand controllers as well as our methods for providing force and haptic feedback are provided in the following.

\subsection{Arm Control \& Feedback}
\label{sec:arm_controller}

The avatar system utilizes two different arm controllers (operator station \& avatar robot), which send joint torque commands to the corresponding Panda arms.
Any information exchanged by both controllers (goal poses and force-torque measurements) are first transformed into a common frame located in the palm of the hands (see \cref{fig:arm_configuration}).
This allows different kinematic chains for the avatar and operator station without any specific retargeting.
The control loop of the whole system runs at 1\,kHz.
At the operator station, the arm controller serves several purposes: First, it measures the 6D human hand pose and sends it to the avatar robot. 
Next, interaction forces measured on the avatar side are displayed to the operator.
All hand movements are measured by the force-torque sensor mounted between the hand exoskeleton and the Panda arm.
The controller uses these measurements to guide the arm following the human movement, generating a weightless feeling for the operator when no force feedback is displayed. Finally, the arm controller pushes the Panda arm away from any joint position or velocity limits.
This is important to prevent the Panda arms from deactivating themselves, as humans can move their arm faster and have a larger workspace.
Avoiding these limits on the operator side is straightforward.
However, limiting the operator input to avoid joint limits on the avatar side, considering different kinematic chains, is not trivial due to latency constraints.
Therefore, we implemented a model-based predictive limit avoidance module that predicts the avatar arm movements based on the current joint state and the target pose commanded by the operator. This way, we can avoid joint limits from the avatar side by displaying forces to the operator.
In addition, in case the operator overcomes the limit forces, the Panda arm will stop (see \cref{sec:safety}) and will be safely restarted (see \cref{sec:robustness}).

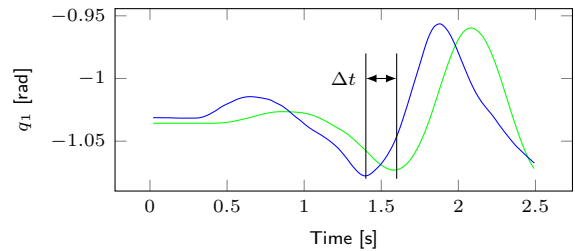
\begin{figure}[b]
\begin{maybepreview}%
  \centering
\pgfplotstableread{data/exp/feedback1.txt}\feedbacktable
\pgfplotstableread{data/exp/model1.txt}\modeltable
  \begin{tikzpicture}[font=\footnotesize\sffamily]
   \begin{axis}[height=4cm,width=\linewidth,
     xlabel={Time [s]}, ylabel={$q_1$ [rad]},]
     \addplot+ [mark=none, green, each nth point=10, filter discard warning=false, unbounded coords=discard] table [x=Time,y=j1] {\feedbacktable};
     \addplot+ [mark=none, blue, each nth point=10, filter discard warning=false, unbounded coords=discard] table [x=Time,y=j1] {\modeltable};
     \draw [black] (1.4, -1.08) -- ( 1.4, -0.98);
     \draw [black] (1.6, -1.08) -- ( 1.6, -0.98);
     \draw[latex-latex] (1.4,-1.0) node [anchor=east] {$\Delta t$} -- (1.6,-1.0) ;
   \end{axis}
  \end{tikzpicture} 
   \end{maybepreview}%
  \caption{Predictive avatar model: Measured joint position for the first joint of the right avatar arm during a grasping motion (green) and predicted
  joint position for predictive limit avoidance (blue). Both measurements are captured on the operator side.
  Communication between both systems and motion execution generate a delay of up to 200\,ms ($\Delta t$), which
  is compensated by the predictive model.}
  \label{fig:exp_model}
\end{figure}

Besides our evaluation at the ANA Avatar XPRIZE Competition (see \cref{sec:competition}), we evaluated subcomponents of our telemanipulation arm controller.
\cref{fig:exp_model} shows the measured and predicted joint position of the first right arm joint during a teleoperated grasping motion.
The prediction compensates the delay, which allows for instantaneous feedback of the avatar arm limits to the operator.
We refer to \citet{lenz2023bimanual} for more details about the telemanipulation controller and component-wise evaluation.

\subsection{Hand Control \& Haptics}
\label{sec:hand_controller}

\begin{figure}
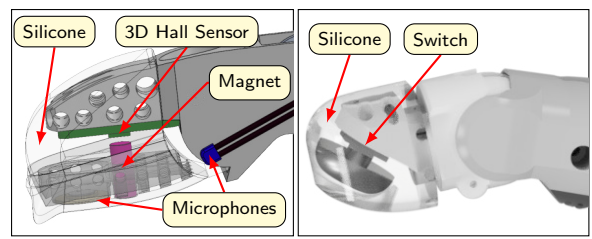

 \centering\begin{maybepreview}%
\begin{tikzpicture}[
  n/.style={fill=yellow!20,rounded corners,draw=black,text=black,align=center,thin,font=\sffamily\footnotesize},
  lab/.style={draw=red,latex-,thick}
]
\node[inner sep=0pt,draw] (sih) {\includegraphics[height=3cm,clip,trim=220px 100px 230px 200px]{images/telemanipulation/sih_cad.PNG}};
\node[inner sep=0pt,draw,right=0.05cm of sih] (svh) {\includegraphics[height=3cm,clip,trim=0px 250px 500px 0px]{images/telemanipulation/svh_cad.PNG}};

\draw[lab] (rel cs:x=0.7,y=0.36,name=sih) -- ++(0.2,-0.5) node[n,anchor=north] (mic) {Microphones};
\coordinate (smic) at (rel cs:x=0.3,y=0.16,name=sih);
\draw[lab,-latex] (mic.west) -- (smic);

\draw[lab] (rel cs:x=0.38,y=0.45,name=sih) -- ++(0.9,1.15) node[n,anchor=south] {3D Hall Sensor};
\draw[lab] (rel cs:x=0.1,y=0.4,name=sih) -- ++(0.2,1.3) node[n,anchor=south] {Silicone};
\draw[lab] (rel cs:x=0.39,y=0.3,name=sih) -- ++(45:1.6cm) -- ++(0.02,0) node[n,anchor=west] {Magnet};

\draw[lab] (rel cs:x=0.1,y=0.5,name=svh) -- ++(0.3,0.9) node[n,anchor=south] {Silicone};
\draw[lab] (rel cs:x=0.23,y=0.4,name=svh) -- ++(1.0,1.2) node[n,anchor=south] {Switch};
\end{tikzpicture}
\end{maybepreview}%
 \caption{Fingertip sensors: SIH (left) and SVH (right).}
 \label{fig:finger_cad}
\end{figure}

The hand controller maps the captured human finger positions to joint position commands for both Schunk hands.
Different mappings are needed for the left and right hand (Schunk SIH and SVH).
The SenseGlove DK1 measures a total of 20\,DoF, 4 joint angles per finger.
In addition, a normalized flexion value is provided, indicating the total flexion of a particular finger.
For the left SIH hand the finger flexion value is used to control the four finger flexion joints (ring finger and pinky are controlled by the same actuator).
The thumb opposition is controlled directly by the corresponding joint measurement from the SenseGlove.
The right Schunk SVH hand has nine actuators and therefore requires more fine-grained joint position commands.
A joint-to-joint mapping is used for the SVH hand using the corresponding SenseGlove joints.
The finger spread is controlled by calculating the angle between the index finger and pinky.

Providing per-finger haptic feedback to the operator when the avatar makes contact with objects or the environment is important for improved telemanipulation.
Both Schunk hands provide motor currents and the measured joint angles.
These measurements can be used to estimate contact or grip forces during active grasping actions.
However, the contact between a finger and the environment when the finger is not actively moving is not visible in the data provided, because most of the finger joints are not backdrivable (SVH hand) or underactuated (SIH hand).
To overcome the lack of information, we designed custom fingertips with additional sensors to replace the original ones.

Each fingertip (except the thumb due to space constrains) on the left SIH hand is equipped with an Adafruit TLV493D 3D magnet hall sensor and a small magnet embedded in a flexible silicon layer (see \cref{fig:finger_cad,fig:roughness_hardware}).
Any contact acting on the fingertip moves the magnet and thus changes the magnetic field measured by the hall sensor.
All hall sensor measurements are collected by a XIAO RP2040 microcontroller with a rate of 400\,Hz and sent to the control PC.
The SenseGlove vibration actuator is triggered for 200\,ms if the absolute magnet field deformation exceeds a predefined threshold.
This gives the operator a brief haptic feedback whenever the fingertip measures any contact.

The right Schunk SVH fingertips do not support the integration of similar Hall sensors due to their small size.
Instead, we integrated 3D pushbutton switches providing binary feedback (see \cref{fig:svh,fig:finger_cad}).
Again, the SenseGlove vibration actuator is triggered for 200\,ms when the contact switch is activated.

The SenseGlove contains active brakes that can prevent the operator from closing a particular finger.
We activate the brake when the motor current from the corresponding avatar finger exceeds a predefined threshold.
Both feedback modalities contribute to an immersive feeling when manipulating in the remote environment.

The ANA Avatar XPRIZE finals required the operator to distinguish different textures by touch alone.
Besides a specific roughness sensing solution described below, we also replaced the left index finger brake
with a Faulhaber LM2070 linear actuator (see \cref{fig:roughness_hardware}).
The actuator pulls the string connected to the SenseGlove fingertip link, enabling the system to actively extend the operator's index finger.
This allows the operator to feel contact forces even when the index finger is not actively moved.

\subsection{Roughness Sensing \& Display}
\label{sec:roughness}

\begin{figure}
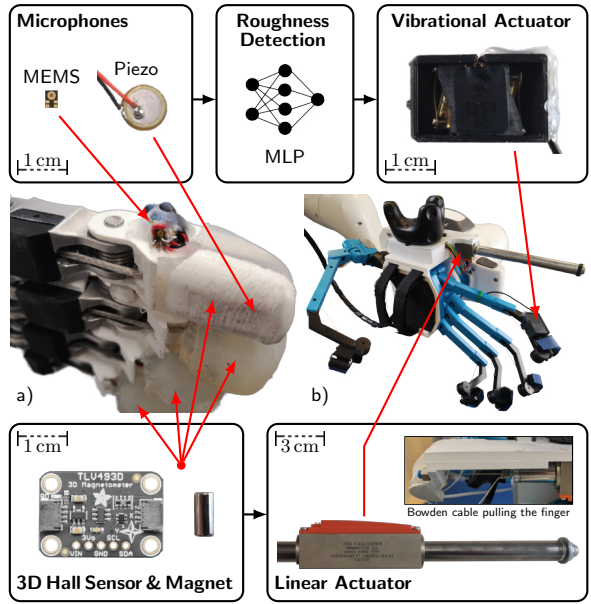

	\centering\begin{maybepreview}%
\begin{tikzpicture}[a/.style={inner sep=0pt}, l/.style={anchor=south west, font=\sffamily\scriptsize, scale=\textscale}, b/.style={anchor=south west, draw, rounded corners, align=center, fill=white, font=\sffamily\scriptsize, scale=\textscale}]
	  \def\fingerheight{.38\linewidth}
	  \definecolor{background1}{rgb}    {0.0,0.0,0.0}
	  \definecolor{background2}{rgb}    {0.39,0.73,0.83}
	  \definecolor{orange}{rgb}         {1.00,0.0,0.00}
	  
	  \def\boxhgap{0.32};
	  \def\boxvgap{0.15};
	  \def\boxheight{2.5};
	  \def\arrowheight{1.25};    
	  \def\boxalen{2.4};
	  \def\boxblen{1.8};
	  
	  \def\boxclen{3.07};
	  \def\boxcheight{2.5};
	  \def\arrowcheight{1.35};

	  \def\textscale{0.8};
	
	  \node[a] (img1) {\includegraphics[height=\fingerheight]{images/roughness_hardware/finger.png}};
	  \node[a, right=.1cm of img1] (img2) {\includegraphics[height=\fingerheight]{images/roughness_hardware/glove2.png}};
	  \node[l] at ($(img1.south west)-(0.01,0.01)$) {a)};
	  \node[l] at ($(img2.south west)-(0.01,0.01)$) {b)};
  
	  \draw[thick, background1, rounded corners] ($(img1.north west)+(0.0,\boxvgap)$) -- ($(img1.north west)+(\boxalen,\boxvgap)$) -- ($(img1.north west)+(\boxalen,\boxheight)$) -- ($(img1.north west)+(0.0,\boxheight)$) -- cycle;
	  \node[anchor = north west, font=\sffamily\scriptsize, scale=\textscale] at ($(img1.north west)+(0.0,\boxheight)-(0.01,0.01)$) {\textbf{Microphones}};
  
	  \draw[thick, background1, rounded corners] ($(img1.north west)+(\boxalen+\boxhgap,\boxvgap)$) -- ($(img1.north west)+(\boxalen+\boxhgap+\boxblen,\boxvgap)$) -- ($(img1.north west)+(\boxalen+\boxhgap+\boxblen,\boxheight)$) -- ($(img1.north west)+(\boxalen+\boxhgap,\boxheight)$) -- cycle;
	  \node[anchor = north west, align=center ,font=\sffamily\scriptsize, scale=\textscale] at ($(img1.north west)+(0.17+\boxalen+\boxhgap,\boxheight)-(0.01,0.01)$) {\textbf{Roughness}\\\textbf{Detection}};
  
	  \draw[thick, background1, rounded corners] ($(img1.north west)+(\boxalen+\boxhgap+\boxblen+\boxhgap,\boxvgap)$) -- ($(img2.north east)+(0.0,\boxvgap)$) -- ($(img2.north east)+(0.0,\boxheight)$) -- ($(img1.north west)+(\boxalen+\boxhgap+\boxblen+\boxhgap,\boxheight)$) -- cycle;
	  \node[anchor = north west, font=\sffamily\scriptsize, scale=\textscale] at ($(img1.north west)+(\boxalen+\boxhgap+\boxblen+\boxhgap,\boxheight)-(-0.08,0.01)$) {\textbf{Vibrational Actuator}};
																																						   			
	  \draw[thick, background1, rounded corners] ($(img1.south west)-(0.0,\boxvgap)$) -- ($(img1.south west)+(\boxclen,-\boxvgap)$) -- ($(img1.south west)+(\boxclen,-\boxcheight)$) -- ($(img1.south west)-(0.0,\boxcheight)$) -- cycle;
	  \node[anchor = south west, font=\sffamily\scriptsize, scale=\textscale] at ($(img1.south west)-(0.0,\boxcheight)-(0.01,-0.01)$) {\textbf{3D\,Hall\,Sensor\,\&\,Magnet}};

	  \draw[thick, background1, rounded corners] ($(img1.south west)+(\boxclen+\boxhgap,-\boxvgap)$) -- ($(img2.south east)+(0.0,-\boxvgap)$) -- ($(img2.south east)+(0.0,-\boxcheight)$) -- ($(img1.south west)+(\boxclen+\boxhgap,-\boxcheight)$) -- cycle;
	  \node[anchor = south west, font=\sffamily\scriptsize, scale=\textscale] at ($(img1.south west)+(\boxclen+\boxhgap,-\boxcheight)-(0.01,-0.06)$) {\textbf{Linear Actuator}};

	  \draw[thick,-latex] ($(img1.north west)+(\boxalen,\arrowheight)$) -- ($(img1.north west)+(\boxalen+\boxhgap,\arrowheight)$);
	  \draw[thick,-latex] ($(img1.north west)+(\boxalen+\boxhgap+\boxblen,\arrowheight)$) -- ($(img1.north west)+(\boxalen+\boxhgap+\boxblen+\boxhgap,\arrowheight)$);
	  \draw[thick,-latex] ($(img1.south west)+(\boxclen,-\arrowcheight)$) -- ($(img1.south west)+(\boxclen+\boxhgap,-\arrowcheight)$);

	  \node(h1)[a, anchor=center] at ($(img1.north west)+(0.55,\arrowheight)$) {\includegraphics[scale=0.08]{images/roughness_hardware/hardware_MEMS.png}};
	  \node[a, right=0.5cm of h1] (h2) {\includegraphics[scale=0.08]{images/roughness_hardware/hardware_piezo.png}};
	  \node(h4)[a, anchor=east] at ($(img2.north east)+(-0.32,1.25)$) {\includegraphics[scale=0.06]{images/roughness_hardware/hardware_basslet.png}};
	  \node(h5)[a, anchor=center] at ($(img1.south west)+(1.2,-\arrowcheight)$) {\includegraphics[scale=0.25]{images/roughness_hardware/magnetometer.png}};
	  \node[a, right=0.4cm of h5] (h6) {\includegraphics[scale=0.08]{images/roughness_hardware/magnet.png}};
	  \node(h7)[a, anchor=west] at ($(img1.south west)+(\boxclen+\boxhgap+0.13,-\arrowcheight-0.4)$) {\includegraphics[scale=0.055]{images/roughness_hardware/faulhaber_crop.png}};

	  \node(h8)[inner sep=0.2pt, draw=black, thick, anchor=west] at ($(img1.south west)+(\boxclen+\boxhgap+1.8,-\arrowcheight+0.583)$) {\includegraphics[scale=0.11]{images/roughness_hardware/faulhaber_thumbnail.png}};

	  \node(scale_a)[a, anchor=south west] at ($(img1.north west)+(0.0,\boxvgap)+(0.1,0.1)$) {\includegraphics[scale=0.08]{images/roughness_hardware/hardware_10mm.png}};
	  \node[font=\sffamily\scriptsize, scale=\textscale, above left=-0.18cm and -0.68cm of scale_a] {$\SI{1}{\cm}$};
  
	  \node(scale_b)[a, anchor=south west] at ($(img1.north west)+(\boxalen+\boxhgap+\boxblen+\boxhgap,\boxvgap)+(0.1,0.1)$) {\includegraphics[scale=0.08]{images/roughness_hardware/hardware_10mm.png}};
	  \node[font=\sffamily\scriptsize, scale=\textscale, above left=-0.18cm and -0.68cm of scale_b] {$\SI{1}{\cm}$};
	  
	  \node(scale_c)[a, anchor=north west] at ($(img1.south west)+(0.0,-\boxvgap)+(0.1,-0.1)$) {\includegraphics[scale=0.08,angle=180,origin=c]{images/roughness_hardware/hardware_10mm.png}};
	  \node[font=\sffamily\scriptsize, scale=\textscale, below left=-0.18cm and -0.68cm of scale_c] {$\SI{1}{\cm}$};

	  \node(scale_c)[a, anchor=north west] at ($(img1.south west)+(\boxclen+\boxhgap,-\boxvgap)+(0.1,-0.1)$) {\includegraphics[scale=0.08,angle=180,origin=c]{images/roughness_hardware/hardware_10mm.png}};
	  \node[font=\sffamily\scriptsize, scale=\textscale, below left=-0.18cm and -0.68cm of scale_c] {$\SI{3}{\cm}$};

	  \node(h1l)[anchor=south, font=\sffamily\scriptsize, scale=\textscale] at ($(h1.north)+(0.0,0.0)$) {MEMS};
	  \node(h2l)[anchor=south, font=\sffamily\scriptsize, scale=\textscale] at ($(h2.north)+(0.05,-0.2)$) {Piezo};
  
	  \node(l1) at ($(img1.north west)+(1.95,-0.5)$) {};
	  \draw[-latex,thick,color=orange] ($(h1.south)+(0.1,-0.1)$) -- (l1);

	  \node(l2) at ($(img1.north west)+(3.3,-1.7)$) {};
	  \draw[-latex,thick,color=orange] ($(h2.south)+(0.28,-0.08)$) -- (l2);

	  \node(l3) at ($(img2.north east)-(0.65,1.8)$) {};
	  \draw[-latex,thick,color=orange] ($(h4.south)+(0.4,0.04)$) -- (l3);

	  \node(mid)[inner sep=0pt, draw=red, fill=red, circle, minimum size=.1cm] at ($(img1.south east)-(1.52,0.7)$) {};

	  \draw[-latex,thick,color=orange] (mid) -- ($(img1.north east)-(1.10,1.4)$);
	  \draw[-latex,thick,color=orange] (mid) -- ($(img1.north east)-(0.80,2.2)$);
	  \draw[-latex,thick,color=orange] (mid) -- ($(img1.north east)-(1.60,2.6)$);
	  \draw[-latex,thick,color=orange] (mid) -- ($(img1.north east)-(2.10,2.8)$);

	  \node(l8) at ($(img2.north east)-(1.6,0.65)$) {};

	  \draw[-latex,thick,color=orange] ($(h7.north)+(-0.84,0.05)$) -| node[]{} ++(-0.0,0.93)-- (l8);

	  \node[font=\sffamily\scriptsize, align=center, scale=0.5, below=-0.02cm of h8.south] {Bowden cable pulling the finger};

	  \node(mlp)[a, anchor=center] at ($(img1.north west)+(\boxalen+\boxhgap+0.9,\arrowheight)$) {\includegraphics[scale=0.03]{images/roughness_hardware/MLP.png}};
	  \node[font=\sffamily\scriptsize, scale=\textscale, below=0.1cm of mlp] {MLP};
	  
	\end{tikzpicture} 
 	\end{maybepreview}%
	\caption{Hardware implementation for roughness sensing and haptic feedback. a) Instrumented index finger on Schunk SIH hand. b) Instrumented index finger on SenseGlove DK1 hand exoskeleton.}
	\label{fig:roughness_hardware}
\end{figure}

The index finger of the left Schunk SIH hand and its counterpart on the SenseGlove DK1 exoskeleton are equipped with a sensor and actuator setup (see \cref{fig:roughness_hardware}),
designed to let the operator intuitively discern between contacts with rough and smooth surfaces~\citep{paetzold2023haptics}.
This low-cost and non-invasive approach uses two microphones capturing vibrations in the finger and the air around it, respectively.
Very short sections \mbox{($\sim10$\,ms)} of these audio streams are then classified by a CNN as either rough or smooth.
Finally, the classification results are used to modulate the frequency and amplitude of an oscillator, generating haptic signal that is directly fed to a vibrational actuator, capable of reproducing a spectrum of frequencies and amplitudes with fast response.
We display rough surfaces by a 60\,Hz sine wave with high amplitude and smooth surfaces by a 120\,Hz sine wave with low amplitude.
We note that the real-time nature of this system means that rough patches (bumps) are felt by the operator, while giving
a slight buzzing sensation on other surfaces.
As the approach is fully based on audio hardware, we integrate it with the rest of the audio system (see \cref{sec:audio}).
This integration also allows us to inject the audio captured by the contact microphone inside the finger into the operator headset, invoking a more realistic and complete haptic perception by hearing the scratching sounds as well.

\begin{figure}
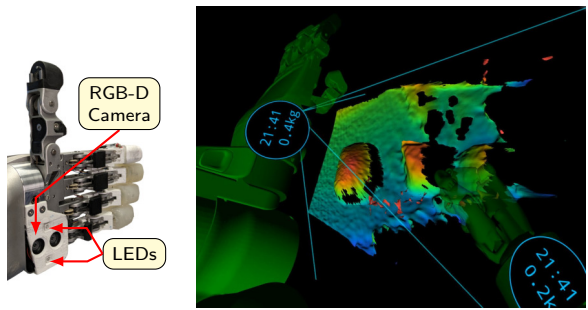

	\centering\begin{maybepreview}%
	\begin{tikzpicture}[
  n/.style={fill=yellow!20,rounded corners,draw=black,text=black,align=center,thin,font=\sffamily\footnotesize},
  lab/.style={draw=red,latex-,thick}
]
\node[inner sep=0pt] (sih) {\includegraphics[height=4cm, clip, trim=160px 30px 150px 50px]{images/telemanipulation/SIH_camera_trans.jpg}};

\draw[lab] (rel cs:x=0.2,y=0.24,name=sih) -- ++ (90:0.5cm) -- ++ (1.1,0.9) node[n,anchor=south] (camera) {RGB-D\\Camera};
\draw[lab] (rel cs:x=0.295,y=0.155,name=sih) -- ++ (0:0.5cm)-- ++(0.2,0.1) node[n,anchor=west] (led) {LEDs};
\coordinate (led2) at (rel cs:x=0.29,y=0.265,name=sih);
\draw[lab,latex-] (led2) -- ++ (0:0.5cm) -- (led.west);
\end{tikzpicture}
	\hfill
	\includegraphics[height=4cm, clip, trim=20px 0px 0px 0px]{images/vr_view/find_stones.jpg}
	\end{maybepreview}%
	\caption{
	Solving the stone task without sight.
	Left: SIH hand equipped with RGB-D camera and LEDs.
	Right: VR visualization.
	 Height is encoded as color (blue to red) and robot arms/hands are shown as a green overlay.
	 The operator can fixate the stone using the right hand, freeze the view using a left thumb gesture,
	 and then touch the stone with the left index finger to feel the texture.
	 }
	\label{fig:vr_view}
\end{figure}

Due to the requirements specified in the competition rules (see \cref{sec:competition}), we designed the system to allow roughness sensing without direct sight.
In this scenario, it is difficult for the operator to even find the objects in order to touch them.
For this reason, we developed a 3D visualization based on geometry captured by a depth camera mounted in the left of the left SIH hand (see \cref{fig:vr_view}).
It allows the operator to locate objects, hold them in place using the right hand, and move the instrumented index finger of the left hand over them.
In the competition, this visualization was not required since the operator could see the stones and was asked to judge whether haptic feedback allowed to discern smooth and rough surfaces.

\subsection{Status Visualization}

\begin{figure}
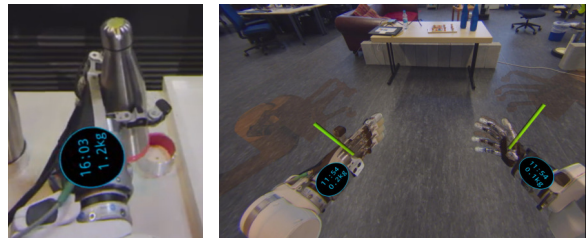

\centering\begin{maybepreview}%
 \includegraphics[height=3.1cm, clip, trim=650px 450px 480px 750px]{images/telemanipulation/wrist_watch.png}\hfill%
 \includegraphics[height=3.1cm, clip, trim=100px 200px 0px 400px]{images/fade_in2.png}
 \end{maybepreview}%
 \caption{Left: Wrist watch VR overlay shows current time and estimated weight.
        Right: If operator and avatar arm poses differ, a fade-in sequence is initiated. Rendered overlays
        show the operator arm pose, i.e. where the avatar arms will move to once the system is activated.}
 \label{fig:VR_overlay}
\end{figure}

Situational awareness for both the remote environment and the system status is important for successful teleoperation.
Most of the developed system components focus on presenting the remote environment as immersive as possible while appealing to multiple human senses.
Providing any additional information such as system health and different sensor measurements should not break the immersive teleoperation experience of the operator.
Therefore, we implemented VR overlays which display additional information in natural ways.
The current time and payload are rendered on each arm as a virtual wrist watch (see \cref{fig:VR_overlay}).

The telemanipulation subsystem is complex and does not always operate as expected by the operator.
Important error notifications are displayed to the operator in various ways:
A pressed E-Stop on the avatar side results in a red view for the operator.
If the operator exceeds the head arm workspace by moving the head too far, the view will fade to black and an error message will be displayed.
Similar, if one of the arms cannot follow the operator's movement due to safety stops, network problems, or simply because the system is not activated, colored arm models are shown (see \cref{fig:VR_overlay}) to indicate
that the system is not following.
To minimize operator distraction during normal operation, system status indicators are not visible when everything is running correctly.

\subsection{Calibration}
\label{sec:calibration}

Before the avatar system can be used, multiple transforms and parameters need to be calibrated.
We devised a principled approach starting at camera intrinsic calibration over hand-eye calibration on the robot side, to VR calibration on the operator side.

Intrinsic camera calibration is done using the kalibr software package~\citep{rehder2016extending}.
The main vision cameras and ground cameras of the avatar robot have very high FoV with significant fish-eye distortion, thus the Double-Sphere camera model~\citep{usenko2018double} is used to describe the intrinsics.

Extrinsic hand-eye calibration estimates the transformations between the cameras, the head arm, and the two main arms~\citep{schwarz2021vr}.
For this purpose, 3D-printed ArUco markers are mounted on the robot wrists.
We use the ArUco marker detector of the OpenCV library to extract 2D pixel coordinates.
During sample collection, the head continuously moves in a predefined sinusoidal pattern while the robot arm is moved manually using teach mode.
Finally, the samples are used to compute optimal transforms using the Ceres solver.

The operator station is calibrated using the VR tracking setup. For this purpose, VR trackers are mounted on the exoskeleton wrists.
After assembly, the arms are moved using teach mode and tracking poses are recorded. This allows precise estimation of the arm mounting
poses and the operator station base pose relative to the VR coordinate system.

In addition to camera and robot calibration, the force-torque sensors at each wrist need to be calibrated~\citep{lenz2023bimanual}.
Different end-effectors (SenseGloves, Schunk SIH, Schunk SVH hand, and corresponding 3D printed mounting adapters) result in different masses and center of mass.
In addition, sensor bias results in barely usable raw sensor data.
For calibration, 20 data samples from different sensor poses are collected.
A standard least squares solver estimates the sensor parameters, i.e. the force and torque bias and the mass and center of mass of all attached components to compensate these effects.
The calibration is performed once after every hardware change at the end-effectors or if the bias drift is too large.
This method does not compensate for bias drift online, but is sufficient for our application.

Finally, the nominal head pose is calibrated once the operator sits comfortably in the chair.

\section{Locomotion}

The omnidirectional base gives the operator advanced maneuverability.
Four mecanum wheels with 8\,inch (20.32\,cm) diameter  are driven by one RMD-X8 brushless motor each.
The motor has a built-in 6.2:1 ratio gearbox and is connected via a 1:1 timing belt.
To indicate the avatar's movement to people in the remote environment, addressable RGB LEDs are mounted under the base plate.
When the avatar is moving, the LEDs in the corresponding direction will light up.
Running lights in clockwise or counter clockwise direction indicate rotation in place.
In addition, the LEDs indicate a pressed E-Stop and the battery level during charging.
A Raspberry Pi 4 Model B is used to communicate with all motors via a CAN interface, commands the avatar's height (see \cref{sec:pedal}), reads the current battery information, and sets the LED colors.
The Raspberry Pi is connected to the main PC via Ethernet and runs a separate ROS instance.
The base can be controlled using a standard wireless Xbox controller independent from the main PC.

\subsection{3D Rudder}

\begin{figure}
\begin{maybepreview}%
\tikzset{
    n/.style={fill=yellow!20,rounded corners,draw=black,text=black,align=center,thin,font=\sffamily\footnotesize},
    lab/.style={draw=red,latex-,thick}
  }
  \centering
  \tikz[]{
    \node[inner sep=0pt] (rudder) {\includegraphics[height=3cm]{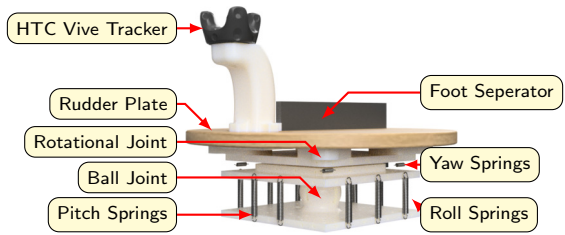}};

    \coordinate (rudderw) at ($(rudder.west)+(-0.1,0)$);

    \draw[lab] (rel cs:x=0.07,y=0.9,name=rudder) -- (\currentcoordinate-|rudderw) node [n,anchor=east] {HTC Vive Tracker};
    \draw[lab] (rel cs:x=0.1,y=0.45,name=rudder) -- ++(135:0.5cm) -- (\currentcoordinate-|rudderw) node [n,anchor=east] {Rudder Plate};
    \draw[lab] (rel cs:x=0.5,y=0.33,name=rudder) -- ++(135:0.3cm) -- (\currentcoordinate-|rudderw) node [n,anchor=east] {Rotational Joint};
    \draw[lab] (rel cs:x=0.5,y=0.15,name=rudder) -- ++(135:0.4cm) -- (\currentcoordinate-|rudderw) node [n,anchor=east] {Ball Joint};
    \draw[lab] (rel cs:x=0.25,y=0.08,name=rudder) -- (\currentcoordinate-|rudderw) node [n,anchor=east] {Pitch Springs};

    \coordinate (ruddere) at ($(rudder.east)+(-0.5,0)$);

    \draw[lab] (rel cs:x=0.5,y=0.5,name=rudder) -- ++(45:0.5cm) -- (\currentcoordinate-|ruddere) node [n,anchor=west] {Foot Seperator};
    \draw[lab] (rel cs:x=0.8,y=0.3,name=rudder) -- (\currentcoordinate-|ruddere) node [n,anchor=west] {Yaw Springs};
    \draw[lab] (rel cs:x=0.83,y=0.15,name=rudder) -- ++(-45:0.3cm) -- (\currentcoordinate-|ruddere) node [n,anchor=west] {Roll Springs};
  } 
   \end{maybepreview}%
  \caption{Self-centering 3D rudder design with individually tunable springs for intuitive locomotion control.}
  \vspace{-1ex}
  \label{fig:rudder}
\end{figure}

Providing locomotion control for a holonomic avatar robot platform in a VR setting is a challenging task. 
In our setup, control methods are constrained as the operator's arms and hands control the bimanual arm and dexterous hands, and the VR headset
pose controls the robot's 6D-movable head. 
We propose a 3D rudder foot input device with individually tunable springs for intuitive locomotion control (see \cref{fig:rudder}).

The springs provide resistance and self-centering of the foot platform.
The mechanical base of the rudder is built around a ball-bearing joint and a rotational thrust-bearing joint.
Springs with different tension allow individual control of the resistance per axis.
For absolute pose estimates, we attach an HTC Vive tracker to the rudder, which receives signals from the VR tracking system. The measured orientation (relative to the start orientation) is then translated to movement commands.
We place a foot separator on the middle of the rudder's surface to ease blind foot placement.
In contrast to commercially available devices, our input device requires no calibration step by the operator.

Using the feet to pitch the rudder results in an intuitive control of the robot base for moving forward and backwards.
Rolling the feet to the left and right allows for sideways control.
Rotating the rudder on the yaw axis results in a rotation of the robot base.
Mechanical end stops on the rotational axis prevent over-bending the springs.

\subsection{Locomotion Visualizations}
\label{sec:birds-eye}

\begin{figure}
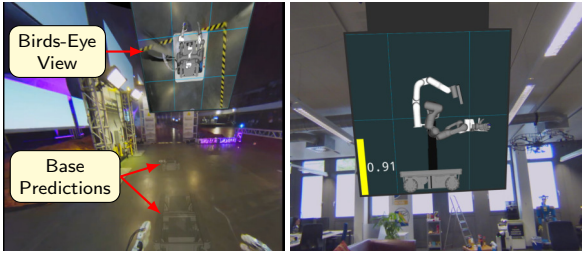

 \centering\begin{maybepreview}%
 \tikz[
    font=\footnotesize\sffamily,
    box/.style={m,draw=black,thin,align=center},
	lab/.style={draw=red,latex-,thick}
   ]{
   \node[inner sep=0cm] (img) {\includegraphics[height=3.3cm,clip,trim=0px 300px 0px 0px]{images/locomotion/birds-eye.png}};
   \draw[lab] (rel cs:x=.5,y=.8,name=img) -- ++(-0.5,0) node [box,anchor=east] {Birds-Eye\\View};

   \node[box] at (rel cs:x=.22,y=0.3,name=img) (l) {Base\\Predictions};
   \draw[lab] (rel cs:x=.56,y=.35,name=img) -- (l.east);
   \draw[lab] (rel cs:x=.56,y=.15,name=img) -- (l.east);
 }\hfill%
 \includegraphics[height=3.3cm,clip,trim=400px 450px 150px 300px]{images/height_adjust.png}
 \end{maybepreview}%
 \caption{Locomotion visualizations. Left: Birds-Eye view and predictions of base pose in the future.
 Right: Side-view for height adjustment.}
 \label{fig:birds_eye}
\end{figure}

Despite the movable wide-angle cameras, the operator has limited view to the side and behind the robot.
To overcome this limitation, especially for situation where the operator drives backwards, a rendered birds-eye view similar to a rearview mirror in a car is used to display additional information.
The view slides down into the field of view when the avatar drives backwards or the operator looks upwards.
Input for this comes from two Logitech Brio webcams with wide-angle converter that are mounted on the avatar's upper body facing to the front and back of the robot. \cref{fig:sys_overview} shows the location of the front camera. The rear camera is mounted in a similar way.
The video streams are projected onto the ground plane using the camera calibration, stitched together with per-camera alpha masks, and displayed in the birds-eye view (see \cref{fig:birds_eye}). The camera extrinsics were calibrated using a marker pattern laid in the area to the sides of the robot, which is visible in both cameras.
The alpha masks are chosen in such a way that interference from e.g. the robot elbows is minimized. In some cases, parts of the elbow are still visible in the images, but this has not led to confusion of the operators so far. Furthermore, the projection and stitching assumes that all objects are on ground level. Violation of this assumption will lead to stitching errors.
However, the boundary between the floor and any object will always be
at the correct location.
The ``rear mirror'' slides out of view when the operator stops driving and looks below the imaginary horizon for 1\,second.

In addition, a 3D predictive model of the robot's base is rendered while driving to facilitate anticipative navigation.
\cref{fig:birds_eye} shows this base model in the VR view. The models display the base location in 2.5 and 5\,seconds in the future, assuming constant velocity.

\subsection{Avatar Height Adjustment}
\label{sec:pedal}

The upper body of our avatar robot can be adjusted in height to support manipulation at different heights and communication with standing and sitting persons.
The linear axis adjusting the avatar's height (shoulder heights reaching from 98\,cm to 152\,cm) is controlled using a bi-directional Danfoss KEP foot pedal (see \cref{fig:sys_overview}). 
Tilting it forward will lift the robot, tilting it back will lower the robot.
While adjusting the avatar's height, the operator sees a rendered side view of the robot model, giving a better understanding of the current height (see \cref{fig:birds_eye}).

\section{Wireless Communication}
\label{sec:network}

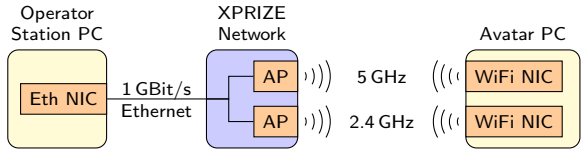
\begin{figure}
 \centering\begin{maybepreview}%
\begin{tikzpicture}[transmission/.style={decorate, decoration={expanding waves, angle=30,
                          segment length=3}}]
  \node[m,minimum width=1.3cm,minimum height=1.3cm,label={[align=center]north:Operator\\Station PC}] (op) at (0,0) {};
  \node[anchor=east,draw,fill=orange!40] (eth) at (op.east) {Eth NIC};

  \node[m,fill=blue!20,minimum width=1.2cm,minimum height=1.3cm,right=1.3cm of op,label={[align=center]north:XPRIZE\\Network}] (xprize) {};
  \node[anchor=east,draw,fill=orange!40] (ap5) at ($(xprize.east)+(0,+0.3)$) {AP};
  \node[anchor=east,draw,fill=orange!40] (ap2) at ($(xprize.east)+(0,-0.3)$) {AP};

  \node[m,minimum width=1.5cm,minimum height=1.3cm,right=2.2cm of xprize,label={[align=center]north:Avatar PC}] (avatar) {};
  \node[anchor=west,draw,fill=orange!40] (w5) at (ap5-|avatar.west) {WiFi NIC};
  \node[anchor=west,draw,fill=orange!40] (w2) at (ap2-|avatar.west) {WiFi NIC};

  \draw (op) -- (xprize) node[midway,align=center,font=\sffamily\footnotesize] {1\,GBit/s\\Ethernet};

  \node (wifi5) at ($(ap5.east)!0.5!(w5.west)$) {5\,GHz};
  \node (wifi2) at ($(ap2.east)!0.5!(w2.west)$) {2.4\,GHz};

  \draw[transmission] (ap5.east) -- ++(0.5,0);
  \draw[transmission] (w5.west) -- ++(-0.5,0);

  \draw[transmission] (ap2.east) -- ++(0.5,0);
  \draw[transmission] (w2.west) -- ++(-0.5,0);

  \draw (xprize.west) -- ++(0.3,0) |- (ap5);
  \draw (xprize.west) -- ++(0.3,0) |- (ap2);
 \end{tikzpicture} 
  \end{maybepreview}%
 \caption{Network Architecture. The operator station contains a 1\,GBit/s ethernet adapter, which is connected
 to the XPRIZE network (or our own AP during testing). Two separate access points broadcast a WiFi network at
 2.4\,GHz and 5\,GHz, respectively.
 The avatar control PC is equipped with two WiFi adapters.
 Adapted from \citet{schwarz2023robust}.}
 \label{fig:network_architecture}
\end{figure}

It is clear that true
avatar systems require freedom in mobility, unencumbered by cables.
Our communication system makes use of two WiFi channels in the 2.4\,GHz and 5\,GHz bands, respectively
(see \cref{fig:network_architecture}).
This allows to balance bandwidth across the bands and also to transmit some information redundantly,
increasing robustness to WiFi interference.
We note that WiFi routers commonly offer dual-band operation.

The data streams and bandwidths are configured statically so that there are no bandwidth spikes
at runtime which could lead to sudden WiFi saturation.
The most bandwidth-heavy stream is caused by the main cameras on the robot head with
2$\times$2472$\times$2178\,pixels\,@\,46\,Hz.
We use on-robot GPU-accelerated HEVC encoding and decoding
to compress and decompress the data with minimal latency~\citep{schwarz2023robust}.

\begin{table}[b]
 \centering\footnotesize\setlength{\tabcolsep}{3pt}
 \caption{Bandwidth requirements.}\label{tab:downlink}\label{tab:uplink}
 \begin{tabular}{@{}lrc@{\hspace{3pt}}c@{\hspace{9pt}} lrc@{\hspace{3pt}}c@{}}
  \toprule
  \multicolumn{4}{c}{Downlink from avatar} & \multicolumn{4}{c}{Uplink to avatar}\\
  \cmidrule (r{7pt}) {1-4} \cmidrule () {5-8}
  Channel & \rotatebox[origin=c]{90}{MBit/s} & \rotatebox[origin=c]{90}{5\,GHz} & \rotatebox[origin=c]{90}{2.4GHz} & Channel & \rotatebox[origin=c]{90}{MBit/s} & \rotatebox[origin=c]{90}{5\,GHz} & \rotatebox[origin=c]{90}{2.4\,GHz} \\
  \cmidrule (r{7pt}) {1-4} \cmidrule () {5-8}
  Arm feedback            &  8.5 & $\checkmark$ & $\times$     & Arm control             &  4.9 & $\checkmark$ & $\checkmark$ \\
  Transforms         &  4.1 & $\checkmark$ & $\times$     & Transforms         &  1.4 & $\checkmark$ & $\times$ \\
  Main cameras            & 14.7 & $\checkmark$ & $\times$     & Operator face  &  5.7 & $\times$     & $\checkmark$ \\
  Hand camera     &  5.5 & $\times$     & $\checkmark$ & Audio                   &  0.4 & $\checkmark$ & $\checkmark$ \\
  Diagnostics      &  0.4 & $\checkmark$ & $\checkmark$ \\
  Audio                   &  0.4 & $\checkmark$ & $\checkmark$ \\
  \cmidrule (r{7pt}) {1-4} \cmidrule () {5-8}
  Total [MBit/s]                   &              & 28.1 & 6.3 & Total [MBit/s]            &              & 6.7 & 11.0 \\
  \bottomrule
 \end{tabular}
\end{table}

\Cref{tab:downlink} shows the resulting data bandwidths and channel configurations.
Manipulation control and audio data are transmitted redundantly over both channels,
as they are particularly sensitive to packet drops.

\section{System Safety, Monitoring  \&  Robustness}

Safety and system robustness is key when developing robotic avatar systems for both, the operator station and the avatar robot.
Both systems are designed to directly interact with humans: The operator is strapped into the exoskeleton and the avatar robot can touch and physically interact with recipients.
In this section, we describe our safety measures, our monitoring tools that give the support crew situational awareness over the whole system at one glance, and system robustness procedures.

\subsection{Safety Measures}
\label{sec:safety}

Safety is the number one priority when building a robotic system that will be used by humans to interact with other humans.
We use Franka Emika Panda cobot arms that are designed to work close to people.
Their joint torque measurements per motor allow the controller to stop the arm immediately if an unexpected behavior occurs.

Two different E-Stops are integrated for both, the operator station and the avatar robot.
On the operator side, the software E-Stop stops the arm controller and puts the Panda arms in teach mode.
All motors will hold their position and the support crew can manually move the arm by pressing the teach buttons.
A second E-Stop cuts the power to both arms, causing mechanical brakes to hold each joint in place.

The same two E-Stops are integrated on the avatar robot.
An HRI Wireless Emergency Stop serves as the software stop that puts the Panda arms in teach mode holding their current position.
In addition, the head arm holds the current position and the avatar's base is depowered, allowing a human to push the robot around.
The hardware E-Stop is mounted on the avatar itself and cuts the battery power, resulting in a shutdown of the whole system including all motors, sensors, and the control PC.

\subsection{System Monitoring}
\label{sec:sysmon}

\begin{figure}
 \centering\begin{maybepreview}%
 \includegraphics[width=\linewidth,clip,trim=10px 10px 10px 10px]{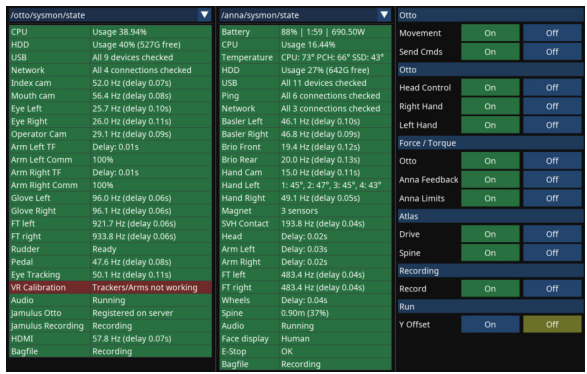}
 \end{maybepreview}%
 \caption{System Monitoring GUI. Left: Operator Station status. Each line corresponds to a system check.
 The red check indicates a problem with the VR trackers mounted on the exoskeleton---caused by
 a support crew member occluding the line-of-sight.
 Center: Avatar robot status.
 Right: Control buttons that enable/disable individual system components.}
 \vspace{-1ex}
 \label{fig:sysmon}
\end{figure}

Monitoring is an essential part of robust robotics. It allows engineers to analyze problems and find their
causes quickly.
In our scenario, it was especially important to make sure the system is healthy before starting a run, since
from then on, manual intervention was not permitted. During the run, the role of monitoring switches to
a safety perspective, allowing the support crew to abort the run in case of danger to the human operator, the robot,
or the environment.
To be able to monitor the highly complex avatar system with one glance, we developed an integrated GUI.
Because it contains a multitude of video streams and complex plots, the standard ROS GUI, \texttt{rqt}, was not
suitable, as it is not optimized for high-bandwidth display.
Instead, we developed a GUI based on \texttt{imgui}\footnote{\url{https://github.com/ocornut/imgui}}, an
immediate-mode GUI toolkit with OpenGL bindings. This allows us to decode and display the video streams
directly on the GPU. The GUI follows the \texttt{rqt} paradigm with individual widgets that can be arranged via
drag \& drop.

The most important monitoring display is shown in \cref{fig:sysmon}. Both operator station and avatar robot
run a \texttt{sysmon} node, which performs several checks with 1\,Hz.
These checks range from ``Is hardware device X connected?'' over ``Does component Y produce data?'' to ``Is the operator station properly calibrated?''.
The intention is simple: If all checks are successful, the support crew can start the run with confidence.
Indeed, our policy was that every time an undetected error or misconfiguration led to a sub-optimal test run,
a specific check for this condition was added.
Overall, checks are similar to unit tests in software engineering, but monitor the live system in hardware \textit{and} software.

\begin{figure}
 \centering\begin{maybepreview}%
 \includegraphics[width=\linewidth]{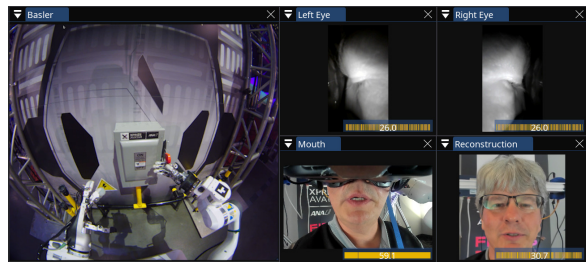}
 \end{maybepreview}%
 \caption{Camera streams. Left: Raw wide-angle camera stream (left eye) from the robot.
 Right: Eye cameras, mouth camera, and reconstructed animated face of the operator.}\label{fig:cams}
\end{figure}

Additionally, a section of the GUI with camera streams (see \cref{fig:cams}) together with headsets providing audio feedback give situational awareness to the
support crew.

\subsection{System Robustness}
\label{sec:robustness}

Ensuring support crew situational awareness and the connectionless network system are features that make
the system more robust. However, there are many problems that can occur during a run, where manual
intervention is not possible without aborting the trial.
For this reason, we added auto-recovery mechanisms on multiple layers.

First, the Franka Emika Panda arms have independent safety systems which detect unsafe situations and
either perform a soft-stop (braking with motor power) or hard-stop (engaging hardware brakes and switching off motor power).
Since the operator can trigger both, e.g. by hitting an object with high speed, it is desirable
to recover from these conditions.
To this end, we modified the Panda firmware to be able to trigger recovery from an autonomous observer, which
restarts the arms as long as the manual E-Stop is not triggered.
During the restart of the arm, the operator is shown a 3D model of the arm to indicate that the arm is restarting
and they should wait until the process is finished. The arm pose is then softly faded to the current operator pose
and operation can continue~\citep{lenz2023bimanual}.

Secondly, many hard- and software problems can be solved by simply restarting the affected processes~\citep{microreboot}.
As a simple example, restarting a device driver ROS node will recover after a transient disconnection of the device,
without the need to make the driver node itself robust against such events.
We stringently use the \texttt{respawn} feature of the ROS launch system to ensure that all nodes are
automatically restarted whenever they exit.
Furthermore, watchdog mechanisms are integrated that force nodes to exit which are stuck
and do not produce output.

Finally, as a last line of defense, the main control PC is equipped with an external watchdog device.
Our software running on the control PC regularly resets this watchdog. Should the system hang completely
(which happened once during testing), the watchdog device will force a reset of the computer.
Consequently, the software is configured to auto-start again, automatically resuming operations.
The complete boot-and-recovery process takes less than one minute.

\section{ANA Avatar XPRIZE Competition}
\label{sec:competition}

The \$10M ANA Avatar XPRIZE competition challenged the robotics community to advance the state of the art in intuitive immersive telepresence systems~\citep{behnke2023avatar}.
The goal was to develop a robotic system that can transport human presence to a remote location in real time.
A total of 99 teams registered in 2019 from 19 different countries across the world.
One focus was pointed to intuitive and easy control of the avatar system.
Thus, a panel of international experts with experience in related research fields were selected to evaluate all proposed systems at the semifinals and finals. In both events, one judge (the operator) controlled the avatar robot in a remote location solving manipulation and locomotion tasks, as well as communicating with a second judge (the recipient) through the avatar system.
In this section, we present quantitative and qualitative results of our very successful participation in the semifinals and finals.

\subsection{Semifinals}

\begin{figure}
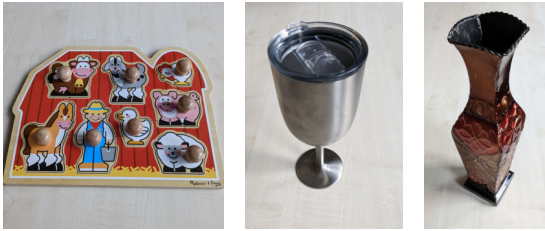

\begin{maybepreview}%
 \hfill%
 \includegraphics[height=3cm, trim={100 550 100 550},clip]{images/semifinal/puzzle.jpg}
 \hfill%
 \includegraphics[height=3cm, trim={400 700 400 100},clip]{images/semifinal/glas.jpg}
 \hfill%
 \includegraphics[height=3cm, trim={500 500 700 100},clip]{images/semifinal/vase.jpg}
 \hfill\strut%
 \end{maybepreview}%
 \caption{Objects used in the ANA Avatar XPRIZE Semifinal scenarios: Solving a jigsaw puzzle (left), celebrating a business deal (middle), and exploring an artifact (right).}
 \label{fig:semifinal_scenarios}
\end{figure}

The semifinals were held in Miami, USA in September 2021 with 29 qualified teams.
An additional 6 teams, unable to travel due to pandemic restrictions, were evaluated in their own labs in early 2022.
All systems were evaluated through three different scenarios: collaboratively solving a puzzle, celebrating a business deal, and exploring an artifact.
The scenario objects are shown in \cref{fig:semifinal_scenarios}.
All scenarios were tested on two days with different operator and recipient judges each.
The best score per scenario over both days was included in the final score.
A maximum of 30\,points were awarded per scenario, based on operator experience (12\,points), recipient experience (8\,points), avatar ability (6\,points), and overall system (4\,points).
In addition, 10\,points were awarded based on a video where teams demonstrated their avatar system with self-chosen tasks in their own lab, resulting in a maximum score of 100\,points.

The operator judge was located in the operator control room, separate from the scenario room where the avatar and recipient judge were located during the test runs.
All communication between the two rooms had to go through the avatar system.
Both the operator station and the avatar robot were allowed to be connected to wired network and power outlets.
Teams had 60\,min to train the operator. The operator then had up to 60\,min to solve all scenarios.
\cref{tab:semifinal_result} shows the semifinal results for the top 20 teams qualified for the finals. Our system was ranked first with 99/100\,points, only missing one point from the recipient experience of Scenario 1. We refer to \citet{lenz2023bimanual} for more in-depth analysis of our semifinal results.

\begin{table}
\centering \footnotesize \setlength\tabcolsep{4pt}
\caption{Results: ANA Avatar XPRIZE semifinals.}
\label{tab:semifinal_result}
\begin{tabular}{@{}r@{\hspace{4pt}}l@{\hspace{2em}}ccc@{\hspace{2em}}c@{\hspace{0.5em}}}\toprule
    &  & \multicolumn{3}{c@{\hspace{2.5em}}}{Scenario}\\
    \cmidrule(r{2em}){3-5}
    Rank & Team\tnote{1} & 1 & 2 & 3 & Total\tnote{2}\\ \midrule
    \textbb{1} & \textbb{NimbRo} & \textbb{29.00} & \textbb{30.00} & \textbb{30.00} & \textbb{99.00}\\
    2 & iCub~\citep{dafarra2022icub3}         & \textbb{29.00} & 28.75 & 27.50 & 95.25\\
    3 & i-Botics~\citep{van2022comes}         & 28.25 & 28.00 & 27.75 & 94.00\\
    4 & Northeastern~\citep{luo2023team}      & 27.50 & 27.75 & 27.75 & 93.00\\
    5 & Dragon Tree Labs                      & 26.75 & 28.50 & 27.75 & 93.00\\
    6 & AVATRINA~\citep{marques2022commodity} & 26.50 & 28.00 & 28.25 & 92.75\\
    7 & Avatar-Hubo~\citep{vaz2022immersive}  & 27.25 & 28.50 & 26.25 & 92.00\\
    8 & Converge                              & 28.25 & 27.00 & 26.75 & 92.00\\
    9 & AlterEgo~\citep{lentini2019alter}     & 27.50 & 26.75 & 27.50 & 91.75\\
    10 & Cyberselves                          & 26.25 & 28.00 & 26.50 & 90.75\\
    11 & Team SNU~\citep{park2022team}        & 27.00 & 27.00 & 25.50 & 89.50\\
    12 & Pollen Robotics                      & 25.75 & 27.75 & 26.00 & 89.50\\
    13 & Last Mile~\citep{lastmile}           & 24.75 & 27.00 & 26.75 & 88.50\\
    14 & Enzo                                 & 25.00 & 27.00 & 25.25 & 87.25\\
    15 & Team UNIST                           & 24.25 & 26.00 & 25.75 & 86.00\\
    16 & Inbiodroid                           & 23.00 & 26.75 & 24.75 & 84.50\\
    17 & Rezilient                            & 24.75 & 24.75 & 24.50 & 84.00\\
    18 & Touchlab                             & 25.25 & 24.50 & 22.75 & 82.50\\
    19 & AvaDynamics                          & 24.75 & 25.50 & 20.25 & 80.50\\
    20 & Janus                                & 21.50 & 24.50 & 24.00 & 80.00\\
    \bottomrule
\end{tabular}
\footnotetext[1]{Only the top 20 teams qualified for the finals are listed.}
\footnotetext[2]{Including 10 video submission points for each team.} 
 \end{table}

\subsection{Finals}

\begin{figure*}
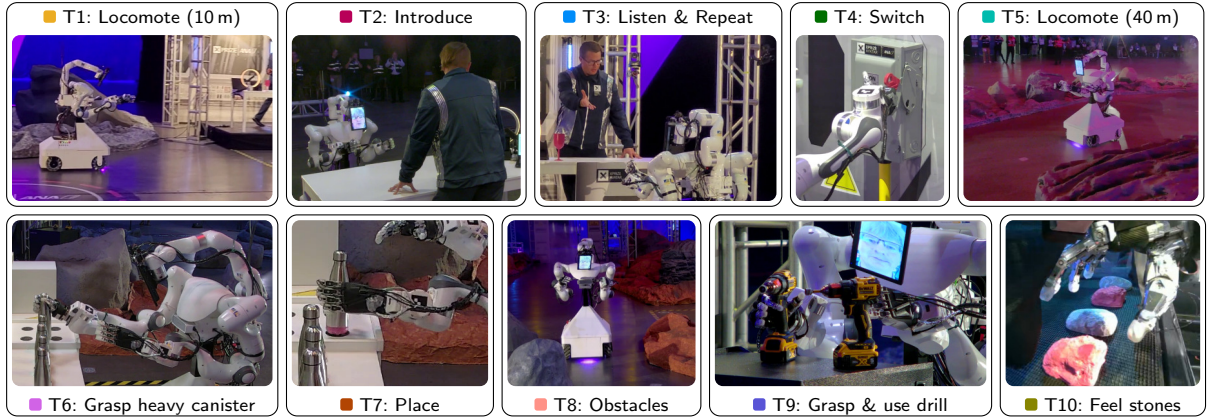

 \newlength{\theight}\setlength\theight{2.2cm}
 \centering\begin{maybepreview}%
 \newcommand{\trbox}[1]{\tikz{\node[inner sep=0pt,rounded corners=3pt,clip]{#1};}}
 \begin{tikzpicture}[
  t/.style={align=center,label distance=2pt, node distance=0.1cm, draw, rounded corners, text depth=0pt, inner sep=2pt}
 ]
 \node[t] (t1) at (-8,0.3) {\taskb{t1} T1: Locomote (10\,m)\\[.4ex]    \trbox{\includegraphics[height=\theight,clip,trim=450px 350px 800px 300px]{images/eval/tasks/t1.png}}};
  \node[t,right=of t1] (t2) {\taskb{t2} T2: Introduce\\[.4ex]          \trbox{\includegraphics[height=\theight,clip,trim=670px 230px 360px 200px]{images/eval/tasks/t2.png}}};
  \node[t,right=of t2] (t3) {\taskb{t3} T3: Listen \& Repeat \\[.4ex]  \trbox{\includegraphics[height=\theight,clip,trim=400px 500px 700px 0px]{images/eval/tasks/t3.png}}};
  \node[t,right=of t3] (t4) {\taskb{t4} T4: Switch\\[.4ex]             \trbox{\includegraphics[height=\theight,clip,trim=1100px 270px 350px 280px]{images/eval/tasks/t4.png}}};
  \node[t,right=of t4] (t5) {\taskb{t5} T5: Locomote (40\,m)\\[.4ex]   \trbox{\includegraphics[height=\theight,clip,trim=800px 230px 150px 160px]{images/eval/tasks/t5_oomph.png}}};

  \node[t,anchor=north west,shift={(0,-0.1)}] (t6) at (t1.south west) {\trbox{\includegraphics[height=\theight,clip,trim=1100px 450px 150px 200px]{images/eval/tasks/t6.png}} \\[.4ex] \taskb{t6} T6: Grasp heavy canister};
  \node[t,right=of t6] (t7) {\trbox{\includegraphics[height=\theight,clip,trim=1150px 500px 500px 350px]{images/eval/tasks/t7.png}} \\[.4ex] \taskb{t7} T7: Place};
  \node[t,right=of t7] (t8) {\trbox{\includegraphics[height=\theight,clip,trim=800px 240px 400px 200px]{images/eval/tasks/t8.png}} \\[.4ex] \taskb{t8} T8: Obstacles};
  \node[t,right=of t8] (t9) {\trbox{\includegraphics[height=\theight,clip,trim=700px 240px 100px 150px]{images/eval/tasks/t9_oomph.png}} \\[.4ex] \taskb{t9} T9: Grasp \& use drill};
  \node[t,right=of t9] (t10) {\trbox{\includegraphics[height=\theight,clip,trim=250px 0px 0px 0px]{images/eval/tasks/t10d.png}} \\[.4ex] \taskb{t10} T10: Feel stones};
  \end{tikzpicture} 
   \end{maybepreview}%
  \caption{Tasks of the ANA Avatar XPRIZE Competition finals.
  \taskb{t1} T1: Short locomotion (approx. 10\,m) to the mission control desk.
  \taskb{t2} T2: The operator introduces themselves to the mission commander.
  \taskb{t3} T3: The operator receives mission details and confirms the tasks.
  \taskb{t4} T4: Activate the power switch.
  \taskb{t5} T5: Approx. 40\,m of locomotion.
  \taskb{t6} T6: Select a canister by weight (approx. 1.2\,kg).
  \taskb{t7} T7: Place the canister in the designated slot.
  \taskb{t8} T8: Navigate around obstacles.
  \taskb{t9} T9: Grasp and use the power drill to unscrew the hex bolt.
  \taskb{t10} T10: Select the rough textured stone based on touch and retrieve it.
  }\label{fig:final_tasks}
\end{figure*}

The ANA Avatar XRIZE finals took place in November 2022 in Long Beach, USA.
A total of 17 qualified teams participated at the non-public qualification day.
The top 16 and 12 teams advanced to the public testing on Day~1 and Day~2, respectively.

Similar to the semifinals, the operator controlling the avatar robot was located in the operator control room, which was separate from the arena where the test course was installed.
This time, teams had to set up their operator station before each test in 30\,min.
Teams had 45\,min in the operator control room to train and familiarize the operator with their system.
In contrast to the semifinals, only one test course was available and therefore the operator training with the avatar took place inside the operator control room without the competition objects.
Shortly before the competition run, the avatar robot was moved to the test course inside the arena where teams connected their system through the competition WiFi network provided by XPRIZE.

The operator judge had up to 25\,min to complete the test course consisting of 10 tasks (see \cref{fig:final_tasks}).
The tasks had to be solved in order and included locomotion, communication with the recipient judge located in the arena, activating a power switch, judging the weight of objects, using a power drill, and distinguishing a rough from a smooth textured stone.
Teams were only allowed to interact with the operator judge after explicit requests by the judge.

\paragraph{Final Results}

The avatar systems were scored based on the task completion and the experience of the operator and recipient judges.
Each successfully completed task was worth 1\,point.
The operator judge awarded up to 1\,point each for the feeling of being present in the remote location, the ability to see and hear clearly, and the ease-of-use of the system.
The recipient judge awarded up to 2\,points for the first two criteria.
Therefore, a maximum score of 15\,points could be achieved.
Ties were broken by completion time.
The team's final score was the better of the two competition days.
\cref{tab:final_points} shows the final result for the 12 teams which advanced to Day~2 of the competition.

Our team NimbRo won the competition with a perfect score of 15 points.
Our operator judges from both competition days were able to solve all ten tasks in 8:15\,min and 5:50\,min, respectively.
In addition, our system received 5/5 judge points both days, resulting in two perfect runs with 15/15\,points.
Only Pollen Robotics were able to also receive a perfect score on Day~2 (they got 14,5\,points on Day~1) but their operator needed almost twice as much time (10:50\,min) to solve all ten tasks.
Pollen Robotics' and our system were the only ones solving all ten tasks on both competition days.
Team Northeastern (placed 3rd) and AVATRINA (placed 4th) managed to solve all tasks on Day~2 and Day~1, respectively.
All competing systems allowed the operator to complete the first four tasks on both days and also Task~5 on at least one day.

\begin{table}
 \centering \footnotesize \setlength\tabcolsep{4pt}
 \caption{Results of the ANA Avatar XPRIZE Finals.}\label{tab:final_points}
\begin{tabular}{@{}r@{\hspace{4pt}}l@{\hspace{6pt}}r@{\hspace{4pt}}r@{\hspace{4pt}}r@{\hspace{6pt}}c}
  \toprule
  &  & \multicolumn{3}{@{}c}{Points} & Time\\
  \cmidrule(r){3-5} \cmidrule(r){6-6}
   Rank & Team                                 & Task & Judge & Total & [mm:ss] \\
  \midrule
  \textbb{1} & \textbb{NimbRo}   & \textbb{10}    & \textbb{5.0}   & \textbb{15.0} & \textbb{\phantom{0}5:50}\\
  2 & Pollen Robotics                          & 10    & 5.0   & 15.0  & 10:50\\
  3 & Team Northeastern~\citep{luo2023team}    & 10    & 4.5   & 14.5  & 21:09\\
  4 & AVATRINA~\citep{marques2022commodity}    & 10    & 4.5   & 14.5  & 24:47\\
  5 & i-Botics~\citep{van2022comes}            & 9     & 5.0   & 14.0  & 25:00\\
  6 & Team UNIST                               & 9     & 4.5   & 13.5  & 25:00\\
  7 & Inbiodroid                               & 8     & 5.0   & 13.0  & 25:00\\
  8 & Team SNU~\citep{park2022team}            & 8     & 4.5   & 12.5  & 25:00\\
  9 & AlterEgo~\citep{lentini2019alter}        & 8     & 4.5   & 12.5  & 25:00\\
  10& Dragon Tree Labs                         & 7     & 4.0   & 11.0  & 25:00\\
  11& Avatar-Hubo~\citep{vaz2022immersive}     & 6     & 3.5   & 9.5   & 25:00\\
  12& Last Mile~\citep{lastmile}               & 5     & 4.0   & 9.0   & 25:00\\
  \bottomrule
 \end{tabular}
 Only the top 12 teams advanced to Day 2 are listed.
  \vspace{-1ex}
\end{table}

\paragraph{Task Completion Times}

\pgfplotstableread[col sep=comma]{data/task_timings.txt}{\loadedtable}
\pgfplotstabletranspose[colnames from=Team]\transposedtable\loadedtable
\begin{figure*}
 \centering\begin{maybepreview}%
 \begin{tikzpicture}[
    ppin/.style={pin edge={latex-,red,thick},pin distance=8pt,font=\sffamily\scriptsize,draw=red,rounded corners},
  ]
  \begin{axis}[
      anchor=north,
      height=4.5cm,
      clip=false,
      xmin=0,
      xmax=27,
      width=.85\linewidth,
      nodes near coords xbar stacked configuration/.style={},
      xbar stacked,
      bar width=12pt,
      axis x line*=left,
      axis y line*=left,
      minor x tick num=1,
      xtick align=outside,
      xtick pos=bottom,
      xtick distance=5,
      minor x tick num=4,
      ytick=data,
      yticklabels from table={\transposedtable}{colnames},
      ytick pos=left,
      xlabel={\footnotesize Time [min:sec]},
      x filter/.code={\pgfmathparse{#1/60}},
      xticklabel={ %
        \pgfmathsetmacro\hours{floor(\tick)}%
        \pgfmathsetmacro\minutes{(\tick-\hours)*0.6}%
        \pgfmathprintnumber{\hours}:\pgfmathprintnumber[fixed, fixed zerofill, skip 0.=true, dec sep={}]{\minutes}%
        },
      printtime/.style={%
        point meta=x,
        nodes near coords style={black},
        nodes near coords={
            \pgfkeys{/pgf/fpu=true}
            \pgfmathparse{\pgfplotspointmeta}
            \pgfmathsetmacro\hours{floor(\pgfplotspointmeta)}%
            \pgfmathsetmacro\minutes{(\pgfplotspointmeta-\hours)*0.6}%
            \pgfmathprintnumber{\hours}:\pgfmathprintnumber[fixed, fixed zerofill, skip 0.=true, dec sep={}]{\minutes}%
            \pgfkeys{/pgf/fpu=false}
        }%
      },
      cycle list={
        {draw=none,fill=t1},
        {draw=none,fill=t2},
        {draw=none,fill=t3},
        {draw=none,fill=t4},
        {draw=none,fill=t5},
        {draw=none,fill=t6},
        {draw=none,fill=t7},
        {draw=none,fill=t8},
        {draw=none,fill=t9},
        {draw=none,fill=t10}
      },
      every plot/.style={draw=none}
    ]

   \foreach \i in {1,...,9} {
    \addplot table [x index=\i, y expr=\coordindex] {\transposedtable};
   }
   \addplot+[printtime] table [x index=10, y expr=\coordindex] {\transposedtable};
  \end{axis}
 \end{tikzpicture} 
  \end{maybepreview}%
 \caption{Per-task execution time for the top six competition runs solving all ten tasks. Tasks are color-coded as in \cref{fig:final_tasks}.}%
 \label{fig:task_timing}
\end{figure*}
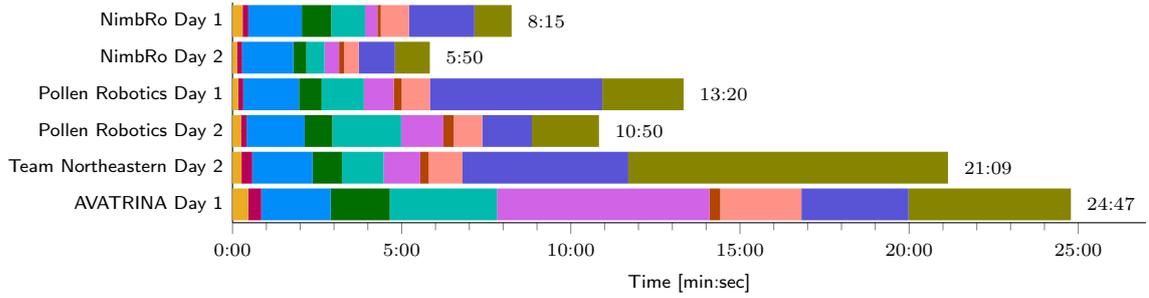

\begin{table*}
 \centering\footnotesize\setlength\tabcolsep{4pt}
\begin{threeparttable}
\newlength{\teamskip}\setlength\teamskip{.12cm}
 \caption{Task completion times at ANA Avatar XPRIZE finals.}\label{tab:nimbro_task_time}
  \begin{tabular}{@{}l@{\hspace{4pt}}c@{\hspace{12pt}}rrrrrrrrrrr@{\hspace{12pt}}r@{}}
 \toprule
    &  & \multicolumn{12}{c}{Time\tnote{1}~~[mm:ss]}\\
  \cmidrule(){3-14}
    Team & Day & Start\tnote{2} & \taskb{t1} T1 & \taskb{t2} T2 & \taskb{t3} T3 & \taskb{t4} T4 & \taskb{t5} T5 & \taskb{t6} T6 & \taskb{t7} T7 & \taskb{t8} T8 & \taskb{t9} T9 & \taskb{t10} T10 & Total\\
    \midrule
    \multirow{3}{*}{NimbRo} & 1 & 00:00 & 00:18 & 00:10 & 01:35 & 00:52 & 01:00 & \textbb{00:22} & \textbb{00:06} & 00:50 & 01:56 & 01:06 & 08:15\\
     & 2 & 00:00 & \textbb{00:08} & \textbb{00:09} & 01:31 & \textbb{00:23} & \textbb{00:32} & 00:26 & 00:09 & \textbb{00:26} & \textbb{01:04} & \textbb{01:02} & \textbb{05:50}\\
     & 1$\rightarrow$2 & \color{teal}{0:00} & \color{teal}{-0:10} & \color{teal}{-0:01} & \color{teal}{-0:04} & \color{teal}{-0:29} & \color{teal}{-0:28} & \color{purple}{+0:04} & \color{purple}{+0:03} & \color{teal}{-0:24} & \color{teal}{-0:52} & \color{teal}{-0:04} & \color{teal}{-2:25}\\[\teamskip]
    \multirow{2}{*}{Pollen Robotics} & 1 & 00:00 & 00:10 & \textbb{00:09} & 01:39 & 00:40 & 01:15 & 00:53 & 00:14 & 00:50 & 05:06 & 02:24 & 13:20\\
     & 2 & 00:00 & 00:15 & \textbb{00:09} & 01:43 & 00:49 & 02:02 & 01:15 & 00:18 & 00:51 & 01:28 & 01:59 & 10:50\\[\teamskip]
    \multirow{2}{*}{Team Northeastern~\citep{luo2023team}} & 1 & 00:00 & 00:33 & 00:24 & 02:08 & 01:43 & 04:03 & 01:27 & 00:36 & 01:56 & &  & 12:50\\
    & 2 & 00:00 & 00:16 & 00:19 & 01:47 & 00:52 & 01:14 & 01:05 & 00:15 & 01:00 & 04:54 & 09:27 & 21:09\\[\teamskip]
    \multirow{2}{*}{AVATRINA~\citep{marques2022commodity}} & 1 & 00:00 & 00:28 & 00:23 & 02:03 & 01:45 & 03:10 & 06:17 & 00:19 & 02:24 & 03:10 & 04:48 & 24:47\\
    & 2 & 00:00 & 00:24 & 00:12 & 01:39 & 01:05 & 02:50 & 00:48 & 00:11 & 01:30 & 02:43 &  & 11:22\\[\teamskip]
    \multirow{2}{*}{i-Botics~\citep{van2022comes}} & 1 & 00:00 & 00:13 & 00:26 & \textbb{01:23} & 01:53 & 01:57 & 01:52 & 02:07 & 02:57 & 09:47 &  & 22:35\\
    & 2 & 00:00 & 00:19 & 00:12 & 01:36 & 03:25 &  &  &  &  &  &  & 05:32\\[\teamskip]
    \multirow{2}{*}{Team UNIST} & 1 & 00:00 & 00:25 & 00:33 & 01:55 & 01:15 & 02:38 & 02:51 & 00:25 & 01:29 & 08:06 &  & 19:37\\
    & 2 & 06:30 & 00:24 & 00:24 & 01:28 & 01:21 & 02:00 & 02:06 & 00:32 & 02:16 & 04:27 &  & 21:28\\[\teamskip]
    \multirow{2}{*}{Inbiodroid} & 1 & 04:46 & 00:23 & 01:44 & 02:05 & 02:28 & 02:26 & 02:03 & 00:44 & 01:26 &  &  & 18:05\\
    & 2 & 05:10 & 00:40 & 00:35 & 02:04 & 03:09 & 02:40 & 04:10 & 00:27 &  &  &  & 18:55\\[\teamskip]
    \multirow{2}{*}{Team SNU~\citep{park2022team}} & 1 & 00:00 & 00:23 & 00:40 & 02:33 & 00:50 & 02:32 & 00:56 & 00:22 & 01:49 &  &  & 10:05\\
    & 2 & 00:00 & 00:26 & 00:18 & 01:36 & 01:26 & 02:39 & 00:32 & 00:30 & 01:49 &  &  & 09:16\\[\teamskip]
    \multirow{2}{*}{AlterEgo~\citep{lentini2019alter}} & 1 & 00:00 & 00:50 & 01:08 & 02:22 & 04:18 & 03:18 & 10:21 &  &  &  &  & 22:17\\
    & 2 & 00:00 & 00:28 & 00:22 & 01:39 & 01:15 & 03:21 & 01:00 & 01:04 & 01:13 &  &  & 10:22\\[\teamskip]
    \multirow{2}{*}{Dragon Tree Labs} & 1 & 05:27 & 00:13 & 00:17 & 01:33 & 05:30 & 01:50 &  &  &  &  &  & 14:50\\
    & 2 & 01:01 & 00:41 & 00:24 & 02:07 & 04:12 & 04:45 & 06:30 & 03:58 &  &  &  & 23:38\\[\teamskip]
    \multirow{2}{*}{Avatar-Hubo~\citep{vaz2022immersive}} & 1 & 00:00 & 00:23 & 00:30 & 01:44 & 00:37 & 07:11 & 03:17 &  &  &  &  & 13:42\\
    & 2 & 00:00 & 00:26 & 00:13 & 01:54 & 02:23 & 02:29 &  &  &  &  &  & 07:25\\[\teamskip]
    \multirow{2}{*}{Last Mile~\citep{lastmile}} & 1 & 00:00 & 03:00 & 02:42 & 01:27 & 03:43 & 04:13 &  &  &  &  &  & 15:05\\
    & 2 & 00:00 & 01:05 & 00:35 & 02:08 & 03:02 & 02:42 &  &  &  &  &  & 09:32\\[\teamskip]
    \midrule
    AvaDynamics\tnote{3} & 1 & 02:50 & 03:19 & 01:06 & 02:11 & 11:39 &  &  &  &  &  &  & 21:05\\[\teamskip]
    iCub~\citep{dafarra2022icub3}\tnote{3} & 1 & 00:53 & 01:09 & 01:12 & 02:08 & 04:28 &  &  &  &  &  &  & 09:50\\[\teamskip]
    Tangible\tnote{3} & 1 & 02:25 & 00:15 & 00:09 & 01:57 &  &  &  &  &  &  &  & 04:46\\[\teamskip]
    Cyberselves -- Touchlab\tnote{3} & 1 & 16:10 & 00:26 & 00:11 & 01:33 & 04:29 &  &  &  &  &  &  & 22:49\\
    \bottomrule
 \end{tabular}
 \begin{tablenotes}
    \item[1]The times were measured from the coverage of the competition: \url{https://www.youtube.com/watch?v=lOnV1Go6Op0}
    \item[2]Time until the avatar started moving.
    \item[3]Team did not advance to Day~2.
\end{tablenotes}

     \vspace{-1ex}
  \end{threeparttable}
\end{table*}

We extracted the per-task completion times for both competition days from the official video feed\footnote{\url{https://www.youtube.com/watch?v=lOnV1Go6Op0}}.
\Cref{tab:nimbro_task_time} reports the task timings for all teams competing at the public competition days. \Cref{fig:task_timing} compares the task timings for the six runs completing all ten tasks.
Both of our competition runs were faster than any other successful run.
As our operator judge on Day~1 solved all tasks in 8:15\,min, giving us a comfortable lead over the other teams,
our operator judge on Day~2 was instructed to take more risks by pushing our system to its limits.
In addition, we greatly increased our avatar's maximum base velocity for Day~2, resulting in much faster execution times
for all tasks involving larger locomotion (Tasks 1, 4, 5, and 8) (see \cref{tab:nimbro_task_time}).
We encountered a minor network issue during Task~9 on Day~1, which explains our longer execution time of 1:56\,min, compared to 1:04\,min on Day~2.
All remaining tasks (2, 3, 6, 7, and 10) were solved within the same time ($\pm$4\,seconds) on both days, showing the robustness of our system.

AVATRINA's system had a much slower drive compared to the top three teams, as evidenced by slower execution times for the locomotion tasks.
The shorter tasks \mbox{1-3} (locomotion and communication with the recipient judge)
and Task~7 (placing the canister into the designated slot) were consistently solved with similar execution times across the top six runs.
Larger differences in individual task execution times are due to subsystem failures or sub-optimal grasp poses in case of the drill (Task~9).
Pollen Robotics' slower locomotion time (Task~5) on Day~2 was due to a reset of the operator control.
AVATRINA had problems during the manipulation in Task~6, which resulted in a software restart on the operator side, costing 2:10\,min.
Both Pollen Robotics on Day~1 and Team Northeastern lost the first drill due to sub-optimal grasp poses.
Both operators had to go back to the table and grasp the second drill before they could complete the task.
Finally, Team Northeastern struggled to reach into the box on Task~10 while grasping the rough stone.
Their avatar's arm kinematics with the wrist above the hand resulted in collisions between the arms and the wall above the box.
The left arm shut down completely during manipulation due to the collision and could not recover.
However, the operator managed to retrieve the correct stone with the right arm after several attempts.

Some teams struggled starting their system in the arena environment using the competition WiFi---resulting in longer waiting times before the avatar began moving (see column ``Start'' in \cref{fig:task_timing}).
For example, the merged team Cyberselves--Touchlab needed over 16\,min to fix and reboot their system, which ended up missing to solve more than four tasks.
This underlines the importance of robustness and ease of operation for such complex systems.

\section{Operator Training}
\label{sec:operator_training}

\begin{figure}
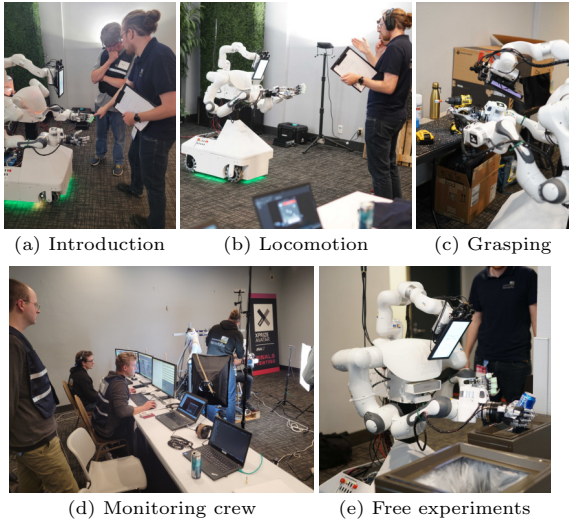

 \centering\footnotesize\begin{maybepreview}%
 \newlength{\opimgh}\setlength{\opimgh}{3cm}
 \setlength{\tabcolsep}{1pt}
 \begin{tabular}{@{}ccc@{}}
 \includegraphics[height=\opimgh, clip, trim=0px 0px 0px 0px]{images/eval/training_person.jpg} &
 \includegraphics[height=\opimgh, clip, trim=400px 0px 0px 0px]{images/eval/training_robot.JPG} &
 \includegraphics[height=\opimgh]{images/eval/training_drill.JPG}\\
 (a) Introduction & (b) Locomotion & (c) Grasping
 \end{tabular}

 \vspace{1ex}

 \begin{tabular}{@{}cc@{}}
 \includegraphics[height=\opimgh, clip, trim=0px 0px 0px 0px]{images/eval/training_crew.jpg} &
 \includegraphics[height=\opimgh, clip, trim=50px 0px 350px 0px]{images/eval/training_can.JPG}\\
 (d) Monitoring crew & (e) Free experiments
 \end{tabular}

 \vspace{1ex}
 \end{maybepreview}%
 \caption{Operator Training. a) Introduction to the avatar robot. b) The operator receives instructions through the system and learns the locomotion capabilities of the system. c) Training to grasp and use the power drill. d) Crew monitors the training and starts individual software components. e) Let the operator playfully enjoy the system by solving competition-unrelated tasks.}
 \label{fig:operator_training}
\end{figure}

In the ANA Avatar XPRIZE Competition, an independent judge acted as the operator, who had never used the system before.
Teams had only 45\,min to train the operator and give them any advice necessary to successfully control the avatar.
Thus, the training had to be optimized to provide enough information without overwhelming the operator.
We present our operator training concept below.

It was crucial to have a clear plan in advance with a defined distribution of work among the team members.
We made this plan about eight weeks before the final event and practiced the training procedure with external, untrained operators in our lab to get enough feedback and resolve missing details.
\Cref{fig:operator_training} shows images from our operator training from the second competition day.

In our team, one dedicated person was responsible for communicating with the judge and managing the training.
This person was the judge's direct contact for any questions.
Next, we had one designated team member who monitored all software components and activated individual modules during the training.
All software components necessary to run the system were started by this person to avoid any miscommunication.
Our monitoring tools (see \cref{sec:sysmon}) were very helpful in providing at-a-glance system status.
These two team members stayed in the operator control room during the competition run and were equipped with headsets that allowed them to listen to the audio communication between the operator and the avatar (see \cref{sec:audio}).
During training and at the specific request of the operator judge, they were able to communicate through this audio channel.
Note that all communications were audible to both the operator and the recipients on the avatar side.
Therefore, the operator support crew could communicate directly with the avatar crew through the system, including the operator judge in any communication.

Next, we had a team member supporting the training by providing all necessary objects and monitoring the hardware of the system (Is the battery charged? Is a cable loose and needs to be fixed?, etc.).
The fourth team member's job was to set up the operator station, including the monitor setup, any cables that needed to be connected, and to launch the team communication components.
The fifth person was responsible for any actions needed in order to start the facial animation (see \cref{sec:facial_animation}), including recording the necessary videos and feeding the data to the algorithm.
The last two people acted as backups and were ready to solve short-term problems.
Otherwise, they kept a low profile to not disrupt the process.

\begin{table}
 \centering\footnotesize\setlength\tabcolsep{4pt}
 \caption{Operator training schedule.}\label{tab:training_schedule}
 \begin{tabular}{lr}
 \toprule
    Training & Time [min]\\
    \midrule
    System overview & 3\\
    Face animation video w/o HMD & 2\\
    Put on HMD & 1\\
    Face animation video with HMD & 2\\
    Strap in hands & 4\\
    Enable arm and hand control & 3\\
    Locomotion training (T1, T5, T8) & 4\\
    Training switch and canister (T4, T6, T7) & 5\\
    Training power drill (T9) & 5\\
    Training stones (T10) & 10\\
    Enjoy the system & 3\\
    System recovery \& recap & 3\\
    \midrule
    Total training & 45\\
    \bottomrule
 \end{tabular}
\end{table}

\begin{figure*}
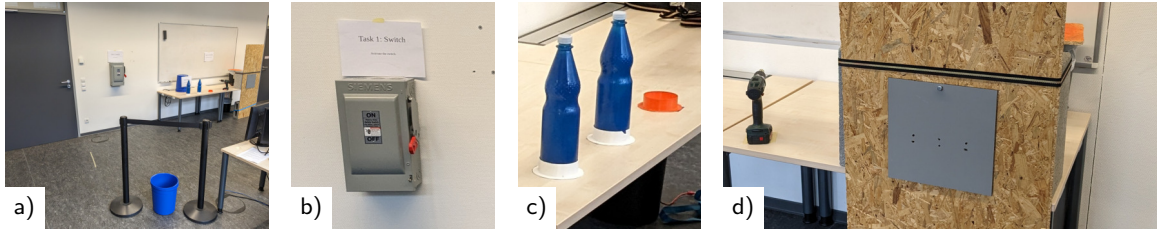

 \centering\begin{maybepreview}%
 \begin{tikzpicture}[a/.style={inner sep=0pt}, l/.style={anchor=south west, fill=white, font=\sffamily\scriptsize}]
    \node[a] (img1) {\includegraphics[height=3cm, clip, trim=0px 0px 550px 0px]{images/userstudy/locomotion.jpg}};
	\node[a, right=.3cm of img1] (img2) {\includegraphics[height=3cm, clip, trim=0px 1200px 3300px 1000px]{images/userstudy/tasks.jpg}};
	\node[a, right=.3cm of img2] (img3) {\includegraphics[height=3cm, clip, trim=1800px 1900px 800px 1600px]{images/userstudy/t2.jpg}};
	\node[a, right=.3cm of img3] (img4) {\includegraphics[height=3cm, clip, trim=1300px 800px 600px 1100px]{images/userstudy/t3.jpg}};

	\node[l] at ($(img1.south west)-(0.01,0.01)$) {a)};
	\node[l] at ($(img2.south west)-(0.01,0.01)$) {b)};
	\node[l] at ($(img3.south west)-(0.01,0.01)$) {c)};
	\node[l] at ($(img4.south west)-(0.01,0.01)$) {d)};
 \end{tikzpicture}
 \end{maybepreview}%
 \caption{User study tasks. a) Test course overview including the barrier that had to be bypassed. b) Activating the switch (\taskb{t4}~T4 in the competition). c) Selecting the heavy bottle and placing it into the orange ring (\taskb{t6}~T6 \& \taskb{t7}~T7). d) Using the power drill to unscrew the hex bolt and opening the hatch (\taskb{t9}~T9).}
 \label{fig:user_study_tasks}
\end{figure*}

Our planned training schedule is summarized in \cref{tab:training_schedule}.
The training started with a brief overview of our system and a short safety briefing.
We wanted the operator to feel safe using our system, which was achieved by providing information about implemented safety features (see \cref{sec:safety}).
The overview eliminated some follow-up questions.
Next, we recorded the videos (with and without HMD) needed for the facial animation method (see \cref{sec:facial_animation}).
The operator had to read a sentence which was placed next to the camera for the first video and displayed in VR for the second video.
While the operator was strapped into the exoskeleton, the HMD displayed the image from the HMD camera facing the room in a look-through mode.
This allowed the operator to see what was happening to their hands.
Strapping the operator's hands into the exoskeleton completed the control preparation, which took about 12\,min.

For the next 30\,min, the operator controlled the avatar and was trained to solve the competition tasks.
First, individual subsystems were activated one at a time: Head movement, arm movement, finger movement, and finally locomotion control.
After activating each subsystem, the operator explored their functionality briefly to get a good system overview (translational head movement to look around objects, birds-eye view for locomotion, force and haptic arm and finger feedback, etc.).
Once the operator was able to control the avatar, the next step was to get used to the easier manipulation tasks (switch and canister).
We used copies of the competition objects to give the operator the best possible training effect.
Until now, the focus had been on the operator discovering the system.
We supported the training by pointing out specific system features, mostly by asking the operator if they noticed them (i.e. Can you feel the force feedback? Can you inspect your hands from different angles?, etc.).
For the more advanced tasks of using the power drill (T9) and feeling the stone texture (T10), we explicitly shared the strategies developed by our expert operators.
Giving the operator the chance to explore different approaches might have led to a better understanding of the system's capabilities, but was not possible due to the very limited training time.

An important aspect of operator training was to make it fun for the operator judge to control the system.
To encourage this, the final training task was always a task not related to the competition.
Examples include the operator throwing away a can or pressing the avatar's own emergency stop.
In addition, we encouraged the operator to look into a mirror and see their own animated face rendered on the avatar screen.

The training ended by unstrapping the operator from the exoskeleton before giving a short summary of the most important points and explaining the system recovery behaviors (see \cref{sec:sysmon}).
The operator had a break of approximately 15\,min while the avatar was moved into the arena.
Right before the run, we strapped the operator back into the system, did final calibrations (head pose and eye tracking), and checked all system components.

All in all, our training preparation has been proven to be successful. Especially the testing of the training procedure in our own lab had solved important details in advance.

\section{User Study}
\label{sec:user_study}

Developing an intuitive telemanipulation and immersive telepresence system for untrained operators was the main target of the ANA Avatar XPRIZE Competition.
After the competition, we evaluated the intuitive control and usability of our system by conducting a user study.
A total of 35 participants with an age of 20 to 34\,years (average of 27.4\,years) operated our avatar system and solved three tasks similar to the competition.
Except for three of our team members, all remaining 32 participants had never controlled our system before.

We divided the participants into three groups: Untrained (18 participants), trained (14), and expert operators (three team members).
All participants started by watching a short introduction video\footnote{\url{https://www.ais.uni-bonn.de/videos/soro_2023_avatar_userstudy/}}, explaining how the avatar system works without giving any hints on solving the tasks.
In addition, the ``trained'' participants were trained for 10\,min to operate the avatar on the test course.
Afterwards, all participants solved the three tasks on the test course using the avatar.
The operator station and avatar robot were located in separate rooms approx. 30\,m apart.

\cref{fig:user_study_tasks} shows the test course and the three tasks in detail.
First, participants had to navigate around the barrier reaching and activating the switch, similar to Task~4 at the competition.
Next, the heavier of two bottles had to be identified and placed inside the orange ring (Tasks 6 \& 7 from the competition).
Both bottles are painted and did not give a visual clue about their weight.
Finally, similar to Task~9 from the competition, participants had to grasp and activate the power drill and use it to unscrew a hex bolt---opening a hatch.
Some participants had difficulties grasping and holding the bottles.
Similar to the competition, where multiple bottles were available, we gave participants multiple tries if they
dropped the bottle out of reach of the avatar (i.e. laying under the table) by resetting the bottle to the
original position.
In case bottles fell over but were still reachable, they were not reset.
The group of participants without training needed bottle resets in 7 out of 18 runs (39\%).
Participants with training dropped the bottles in two cases (14\%).
The expert operators did not drop any bottles.

\newcounter{studyplot}

\newcommand{\plottimes}[1]{%
	\addplot[
		notrain,
		boxplot, boxplot discard if not={Training}{0},
		boxplot/draw position=\thestudyplot
	] table [y={#1},col sep=comma] {data/soro_userstudy_times.txt};
	\addplot[
		gray, opacity=0.5, only marks, discard if not={Training}{0}
	] table [x expr={\thestudyplot},y={#1},col sep=comma] {data/soro_userstudy_times.txt};
	
	\stepcounter{studyplot}

	\addplot[
		train,
		boxplot, boxplot discard if not={Training}{1},
		boxplot/draw position=\thestudyplot
	] table [y={#1},col sep=comma] {data/soro_userstudy_times.txt};
	\addplot[
		gray, opacity=0.5, only marks, discard if not={Training}{1}
	] table [x expr={\thestudyplot},y={#1},col sep=comma] {data/soro_userstudy_times.txt};
	
	\stepcounter{studyplot}

	\addplot[expert,
		boxplot, boxplot discard if not={Training}{2},
		boxplot/draw position=\thestudyplot
	] table [y={#1},col sep=comma] {data/soro_userstudy_times.txt};
	\addplot[
		gray, opacity=0.5, only marks, discard if not={Training}{2}
	] table [x expr={\thestudyplot},y={#1},col sep=comma] {data/soro_userstudy_times.txt};
	
	\stepcounter{studyplot}
}

\begin{figure}
 \centering\begin{maybepreview}%
\pgfplotsset{
    colormap/Set2,
	boxplot/draw direction=y,
	boxplot/hide outliers,
	every axis/.style={scale only axis, mark size=1pt, clip mode=individual},
	notrain/.style={black, boxplot/every box/.style={fill=Set2-A}},
	train/.style={black, boxplot/every box/.style={fill=Set2-B}},
	expert/.style={black, boxplot/every box/.style={fill=Set2-C}}
   }
 \begin{tikzpicture}
  \begin{axis}[
    height=5cm,
    width=1.5cm,
	ymin=0,
	xmin=-1,
	xmax=3,
	xtick pos=bottom,
	xtick={0,1,2},
	xticklabels={Untrained,Trained,Expert},
	x tick label style={rotate=90,anchor=east},
	yticklabel={ %
        \pgfmathsetmacro\minutes{floor(\tick/60)}%
        \pgfmathprintnumber{\minutes}%
    },
    ytick={0,60,...,720},
    title={Total Time},
    ylabel={Time [min]},
    trim axis left
   ]
   
   \setcounter{studyplot}{0}
   \plottimes{time-total}
  \end{axis}
 \end{tikzpicture}\hspace*{0pt}%
 \begin{tikzpicture}
  \begin{axis}[
	height=5cm,
    width=4.5cm,
	ymin=0,
	xmin=-1,
	xmax=11,
	ymax=350,
	xtick pos=bottom,
	xtick={0,1,2,4,5,6,8,9,10},
	xticklabels={Untrained,Trained,Expert,Untrained,Trained,Expert,Untrained,Trained,Expert},
	x tick label style={rotate=90,anchor=east},
	yticklabel={ %
        \pgfmathsetmacro\minutes{floor(\tick/60)}%
        \pgfmathprintnumber{\minutes}%
    },
    ytick={0,60,...,720},
    title={Time Per Task},
    title style={yshift=1ex},
    minor y tick num=1
   ]
   
   \coordinate (top) at (axis description cs:1,1);
   
   \coordinate (l1) at (axis cs:1,0);
   \node[anchor=south] at (l1|-top) {Switch};
   \setcounter{studyplot}{0}
   \plottimes{time-switch}
   
   \stepcounter{studyplot}
   \draw [dashed] (axis cs:3,0) -- (axis cs:3,350);
   
   \coordinate (l2) at (axis cs:5,0);
   \node[anchor=south] at (l2|-top) {Bottles};
   \plottimes{time-bottle}
   
   \stepcounter{studyplot}
   \draw [dashed] (axis cs:7,0) -- (axis cs:7,350);
   
   \coordinate (l2) at (axis cs:9,0);
   \node[anchor=south] at (l2|-top) {Drill};
   \plottimes{time-drill}

  \end{axis}
 \end{tikzpicture} 
  \end{maybepreview}%
 \vspace{-4ex}
 \caption{User study times. Left: Total mission execution time. Right: Times per task.
 Box plots show median and upper/lower quartile. Whiskers show extents disregarding outliers.}
 \label{fig:user_study_times}
\end{figure}
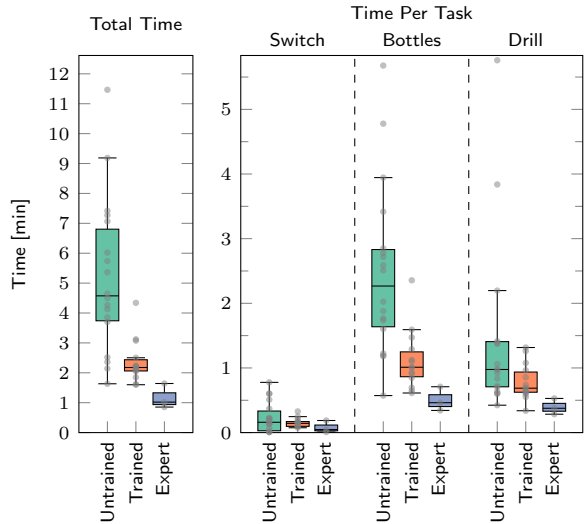

All participants were able to solve all tasks within 12\,min without any help.
\Cref{fig:user_study_times} shows the time needed to solve all tasks and the time per task for all three study groups.
As expected, trained operators were able to solve all tasks on average twice as fast as untrained operators (5:11\,min and 2:22\,min, respectively).
Experts were again twice as fast as trained operators with an average time of 1:10\,min to solve all tasks.
The fact that the execution time for untrained operators only increases by a factor of two underlines the intuitive nature of the control, given the high system complexity.

Feedback from the participants after the study can be summarized as follows:
The vast majority of the participants gave very positive feedback about the immersive and intuitive control of the avatar.
People often forget that their physical body is located in another room while they were controlling the avatar.
Some people had problems with motion sickness.
Most of them had similar experiences with other VR applications or  reading in a car.
The 3D ruder used to control the avatar's locomotion could be improved with stronger springs and a visual overlay showing the current input values to provide a better awareness when people start driving.
In addition, some participants had problems estimating the distance between the avatar and the table:
These operators expected that their legs could collide with the table, since they were sitting. However, the avatar does not have legs.
Clearer visual or auditory cues could resolve this uncertainty for future operators.
Overall, people enjoyed operating our system and felt very present in the remote location.

\section{Lessons Learned}
\label{sec:lessons_learned}

Developing a complex robotic system and especially evaluating the system at a robotic competition gave insights and direct feedback on design choices that we would like to share.

\paragraph{Robustness}
Despite the extensive experience available in our group, the NimbRo avatar system is the most complex system we have ever built.
Good monitoring, failure tolerance, and auto-recovery were extremely important for success.
Our policy of adding a specific system check whenever a hardware or software problem occurred (see \cref{sec:sysmon}) saved a lot of time during development and especially during the competition, when every minute counted.
In contrast to other teams, technical problems did not cause major delays or failures.

\paragraph{Frequent Testing under Competition Conditions}
Having a high test frequency allowed us to identify and address several points that were annoying or uncomfortable
for operators, e.g. resulting in the improved rudder device.
Frequent tests under competition conditions (untrained operator, time and space restrictions, and competition tasks) also helped the support crew to establish routine in efficiently training the operators as evidenced by the detailed routine described in \cref{sec:operator_training}.

\paragraph{1:1 Correspondence is Best}
The connection between operator and avatar needs to be as close to identity as possible.
Avoiding any scaling, offsetting, or 3D processing helps operators to quickly
immerse into the system. In particular, correct hand-eye transformations let operators identify the avatar's hands as their own.

\paragraph{6\,DoF Camera Motion}
The 6D motion of the wide-angle stereo camera mirroring the operator head movement greatly contributed to the immersion in the remote scene and allowed the operator to intuitively chose a viewpoint for manipulation that minimized occlusion.
This especially became evident during the drill task, where operators could look from the side to see the trigger while grasping.

\paragraph{Immersive Control Overlays}
One of our design criteria was to keep the VR overlays minimal and at places which are intuitive for untrained operators (i.e. the wrist watch and birds-eye view as a rear mirror).
Complex overlays and VR menus distract the operator and limit the operator's experience of being present in the remote space.
Most of the operators who have controlled our system have reported immersive telepresence, which we attribute in part to the unobtrusive overlays mentioned above.

\paragraph{Facial Animation and Gestures}
The photorealistic animation of the operator face on the avatar robot together with the realistic display movement and hand gestures enhanced the perception of the operator being present in the avatar robot.

\paragraph{Operators differ}
Humans have different body proportions.
Thus, the operator station needs to support a large variety of different operators (i.e. head size, finger, arm, and leg lengths, etc.).
Our system needs only the initial head reference pose and the input to our face animation pipeline, allowing individual operator to use the system without much preparation or calibration.
While this is true for the vast majority, we have seen problems with very short operators: Reaching the foot pedal and a smaller workspace due to shorter arms were the result.

Over 150 different operators tested our system as part of the user study (\cref{sec:user_study}), during development, or during multiple system demonstrations.
Very few operators complained about motion sickness. Most of them have similar problems with comparable VR applications. 
Individual operators have used the system for over 90\,min at a time with no problems other than mild fatigue.
We conclude, that our system is easy to use even during longer operation sessions, but removing motion sickness for every operator might be impossible.

\paragraph{Modified Components}

Most of the components integrated in our avatar system are off the shelf components.
We spent some effort in exploring the market and finding the best components.
However, some important features were not available on the market.
Therefore, modifying existing components to our needs gave us an advantage over other teams.
Examples include the modified Panda firmware for non-horizontal mounting, additional eye-tracking and mouth cameras on the Valve Index HMD, and modified Schunk hand fingertips with additional contact sensors.

\section{Conclusion}

This article presented the NimbRo avatar system, which won the \$10M ANA Avatar XPRIZE competition.
We describe in detail the avatar robot with a humanoid upper body mounted on a mobile base, and the operator station, which consists of arm and hand exoskeletons, an HMD, and foot pedals.
We provide subsystem evaluations for key components and refer to \citep{schwarz2021nimbro,schwarz2021vr,lenz2023bimanual,schwarz2023robust,paetzold2023haptics,rochow2023attention} for more detailed component analysis.
The robustness of the system, achieved through comprehensive monitoring tools and multi-level system recovery, has been a major contributor to our success.

An important focus of the system design was to provide an intuitive and immersive operator interface for both trained and untrained operators.
Achievement of these design goals is demonstrated in the extensive analysis of the semifinals and finals in the ANA Avatar XPRIZE competition.
Our avatar system allowed the briefly trained judge to complete all ten tasks in 5:50\,min, almost twice as fast as the second placed team.
Key improvements over our semifinal system~\citep{schwarz2021nimbro} such as a new base design, a linear actuator to adjust the torso height, haptic perception, monitoring tools, failure tolerance, and robust wireless communication enabled this success.

Operator training within the limited time frame of 45\,min was one important aspect of the competition.
We described our training approach and team member roles during training in detail.

In addition to the competition analysis, we evaluated our system in a user study.
Untrained operators were able to solve three locomotion and manipulation tasks with and without 10\,min of training in a short time.
Training on the complex system only reduced the execution time by a factor of two, compared to completely untrained operators.
Experts with many hours of experience were only twice as fast as briefly trained operators.
These results underline the intuitiveness and easy usability of the avatar system.

\paragraph{Data availability}
The datasets generated and analysed during this study are available from the corresponding author on reasonable request.

\paragraph{Compliance with Ethical Standards}
Conflict of Interest: The authors declare that they have no conflict of interest.

\printbibliography

\end{document}